%% file: main.tex
\documentclass[a4paper]{article}

\usepackage{afterpage}
\usepackage{algorithm}
\usepackage{algorithmic}
\usepackage[utf8]{inputenc}
\usepackage[T1]{fontenc}
\usepackage[english]{babel}
\usepackage{amssymb}
\usepackage{hyperref}
\usepackage{url}
\usepackage{amsthm}
\usepackage{fullpage}
\usepackage{anyfontsize}
\usepackage{subcaption}
\usepackage{caption}
\usepackage{graphicx}
\usepackage{array,tabularx,booktabs}
\newcolumntype{Y}{>{\raggedright\arraybackslash}X}
\usepackage{amsmath}
\usepackage{olo}
\usepackage[version=4,arrows=pgf-filled,textfontname=sffamily,
mathfontname=mathsf]{mhchem}

\newcommand{\alg}{\textsf{RESPIRE}\xspace}
\newcommand{\PM}{\ce{PM_{2.5}}\xspace}
\newcommand{\RH}{\ce{RH}\xspace}
\newcommand{\CO}{\ce{CO}\xspace}
\newcommand{\Ox}{\ce{O_3}\xspace}
\newcommand{\NOx}{\ce{NO_x}\xspace}
\newcommand{\SOx}{\ce{SO_x}\xspace}
\newcommand{\HT}{\textsf{HT}\xspace}
\newcommand{\OP}{\textsf{OP}\xspace}

\newcommand{\hvy}{\hat\vy}
\newcommand{\vbetao}{\vbeta^\ast}
\newcommand{\veo}{\ve^\ast}
\newcommand{\vyo}{\vy^\ast}
\newcommand{\metric}{$R^2$\xspace}

\newcommand{\mypar}[1]{\noindent\textbf{#1.}}

\begin{document}

\input{frontmatter}

\input{intro}
\input{related}
\input{setting}
\input{method}
\input{exps}

\input{backmatter}

\end{document}

%% file: frontmatter.tex
\title{Provably Outlier-resistant Semi-parametric Regression for Transferable Calibration of Low-cost Air-quality Sensors\thanks{The work was supported by grant no 001296 from the Clean Air Fund.}}

\author{Divyansh Chaurasia\footnote{Indian Institute of Technology, Kanpur} \and Manoj Daram$^{\dagger,}$\footnote{Work done while the author was affiliated to IIT Kanpur} \and Roshan Kumar$^\dagger$ \and Nihal Thukarama Rao$^{\dagger,\ddagger}$ \and Vipul Sangode$^{\dagger,\ddagger}$ \and Pranjal Srivastava$^{\dagger,\ddagger}$ \and Avnish Tripathi$^{\dagger,\ddagger}$ \and Shoubhik Chakraborty\footnote{Shiv Nadar University, Chennai} $^{,\ddagger}$ \and Akanksha$^\dagger$ \and Ambasht Kumar$^\dagger$ \and Davender Sethi$^\dagger$ \and Sachchida Nand Tripathi$^\dagger$ \and Purushottam Kar$^\dagger$}

\date{Correspondence: \url{snt@iitk.ac.in, purushot@cse.iitk.ac.in}}

\maketitle

\begin{abstract}
We present a case study for the calibration of Low-cost air-quality (LCAQ) \CO sensors from one of the largest multi-site-multi-season-multi-sensor-multi-pollutant mobile air-quality monitoring network deployments in India. LCAQ sensors have been shown to play a critical role in the establishment of dense, expansive air-quality monitoring networks and combating elevated pollution levels. The calibration of LCAQ sensors against regulatory-grade monitors is an expensive, laborious and time-consuming process, especially when a large number of sensors are to be deployed in a geographically diverse layout. In this work, we present the \alg technique to calibrate LCAQ sensors to detect ambient \CO (Carbon Monoxide) levels. \alg offers specific advantages over baseline calibration methods popular in literature, such as improved prediction in cross-site, cross-season, and cross-sensor settings. \alg offers a training algorithm that is provably resistant to outliers and an explainable model with the ability to flag instances of model overfitting. Empirical results are presented based on data collected during an extensive deployment spanning four sites, two seasons and six sensor packages. \alg code is available at \url{https://github.com/purushottamkar/respire}.
\end{abstract}

%% file: intro.tex
\section{Introduction}
\label{sec:intro}

Dense air-quality monitoring (AQM) networks play a critical role in the pursuit of sustainable development goals such as Clean Air (UN SDG 3.9). AQM networks enable hotspot detection, source apportionment and can motivate citizen action and regulatory interventions to curb the sources of pollution. The traditional route to establishing AQM networks, often referred to as CAAQMS, uses regulatory-grade instruments (RGI) that pose financial, technical and logistic challenges. CAAQMS stations are expensive, and require specialized personnel during both setup and operation stages. In contrast, low-cost air-quality (LCAQ) sensors have recently emerged as a solution, offering far lower setup costs as well as relatively hands-off operation.

However, LCAQ sensors require on-field calibration against their RGI counterparts as LCAQ sensors can be sensitive to ambient weather conditions and can experience drift across time. The calibration of LCAQ sensors has been a well-studied topic with a variety of machine learning models being used to calibrate LCAQ sensors for pollutants such as \PM, \Ox, \NOx \cite{sirnox,gonzalez2019,spinelle2015,bhowmik2022}. In this work, we present a case study for the calibration of \CO sensors. The data for our study was collected from a multi-site-multi-season-multi-sensor-multi-pollutant mobile air-quality monitoring network established in the city of Lucknow (deployment details in Section~\ref{sec:setting}).\\

\mypar{Our Contributions} Studies exist in literature that attempt to use generic machine learning models such as linear models, trees, neural networks, kernel models to calibrate LCAQ \CO sensors \cite{levy,apostolopoulos,topalovic,zuidema,ariyaratne}. This work identifies specific deficiencies in these approaches and proposes a tailor-made calibration technique called \alg that closely aligns with calibration instructions from the sensor manufacturer. This is done by proposing a semi-parametric regression model instead of a fully non-parametric high capacity model such as decision trees, neural networks, nearest neighbors or kernel regression. \alg also protects the calibration process against outliers by adopting a robust regression technique with provable guarantees. The resulting technique attains a sweet spot in neither overfitting nor underfitting when subjected to data in the wild. Specifically, \alg offers the following advantages:
\begin{enumerate}
    \item \alg offers performance competitive to baseline techniques if tested on data from the training site in the same season. However, \alg offers clear advantages when tested on data from a different site and/or from a different season.
    \item \alg offers provable guarantees of resilience to adversarial outliers.
    \item \alg models lend themselves to compression without sacrificing performance.
    \item A lightweight transfer model is proposed that allows \alg models to not just be used across sites and seasons, but across different sensors as well. This transfer does not require ground truth and can be done in a completely unsupervised manner. This is advantageous for large LCAQ sensor networks, where calibrating each sensor individually is expensive, laborious and time-consuming.
    \item The semi-parametric nature of \alg allows it to flag certain instances of overfitting that allowed identification of cases where readings from different sensors were swapped due to a firmware issue.
\end{enumerate}

\mypar{Key Takeaways} We summarize case study insights using empirical data from Section~\ref{sec:exps},
\begin{enumerate}
    \item Most methods proposed in literature, and indeed generic implementation of popular machine learning models, offer excellent training performance, with train \metric values (the metric being the proportion of explained variance aka \emph{coefficient of determination}) well above 0.95 in most cases. This indicates that most methods have sufficient capacity to explain the data. (fig - ~\ref{fig:training_performance})
    \item Performance drops appreciably even on test data taken from the same site and during the same season. This drop is sharper for high-capacity methods such as random forests and nearest neighbours, indicating a higher degree of overfitting, although all methods notice a drop. (fig - ~\ref{fig:heatmap_ss})
    \item The drop is sharper when tested on the same site but a different season or a different site altogether. This indicates that site-specific characteristics and weather conditions such as temperature and humidity have a significant affect on calibration performance. (fig - ~\ref{fig:heatmap_sx})
    \item When comparing performance across sensors, across sites and across seasons, \alg offers the largest number of wins, followed by linear regression, followed by multi-layered perceptron. Other methods, such as GBDT, kernel regression, and nearest-neighbours, seem excessively prone to overfitting and perform well only if tested on the same site during the same season. (fig - ~\ref{fig:transfer_no_adapt})
    \item A possible cause of performance drop when tested on different site/season seems to be a certain degree of drift in the RGI readings, i.e. the ground truth itself. However, this can be readily corrected by learning a simple 1D adapter. This is suspected because (a) despite lower \metric values, predictions often closely track the ground truth in shape, being off by just a shift or a scale (b) learning a simple 1D adapter model causes significant rise in \metric, and (c) RGI \CO analyzers have to regularly undergo manual calibration and thus, zero-shifts or sensitivity changes are expected. (fig - ~\ref{fig:transfer_adapt})
    \item Another probable cause of performance drop seems to be the presence of outliers. This is indicated by the fact that in several instances, test \metric scores rise rapidly if a small, say 5\%, fraction of test data points are excluded from evaluation. (fig - \ref{fig:row3})
    \item \alg models can be transferred from one LCAQ sensor to another. The transfer performance can be enhanced by learning a self-supervised linear transfer model. (fig - ~\ref{fig:s2s_transfer})
    \item \alg models can be significantly compressed without much loss of performance. (fig - ~\ref{fig:compression_r2_drop})
    \item Their semi-parametric nature allows \alg to flag certain cases of overfitting. (fig - ~\ref{fig:row4})
\end{enumerate}

%% file: related.tex
\section{Related Work}
\label{sec:related}
Low-cost air quality (LCAQ) sensors offer a promising solution for high spatiotemporal resolution air quality monitoring, supplementing traditional regulatory-grade CAAQMS sites \cite{sirnox, zuidema, topalovic}. However, the raw data from LCAQ sensors often suffers from low accuracy, drift, and cross-sensitivities to other pollutants and environmental conditions, necessitating extensive in-field calibration to ensure data reliability \cite{apostolopoulos, karagulian2019}.

Past research has focused on various field calibration methods using both linear and non-linear regression models. While simpler models can be effective for some pollutants like \Ox in certain environments \cite{spinelle2015}, many studies demonstrate the superior performance of non-linear ML approaches. For instance, Topalović et al. \cite{topalovic} found that Artificial Neural Networks (ANNs) yielded better results for calibrating \CO and \Ox sensors compared to linear models. Similarly, Sahu et al. \cite{sirnox} proposed a non-parametric nearest-neighbors-based algorithm that improved the coefficient of determination ($R^2$) metric by 4–20 percentage points for \Ox and \NOx over classical methods. Apostolopoulos et al. \cite{apostolopoulos} also highlighted the promise of the Random Forest algorithm for calibrating LCAQ devices in urban settings.

Multiple studies agree on the importance of incorporating factors such as ambient temperature and relative humidity (\RH) into models to improve calibration accuracy \cite{sirnox, zuidema, ariyaratne}, as is also corroborated by manufacturer instructions. Leveraging sensor cross-sensitivities by including measurements of co-pollutants as predictors can enhance model performance. For example, Topalović et al. \cite{topalovic} identified \NOx and \PM as valuable predictors for \CO calibration. In a novel approach, Apostolopoulos et al. \cite{apostolopoulos} demonstrated that using raw sensor voltage signals directly within an ML algorithm, rather than manufacturer-calibrated values, can reduce inter-unit variability and improve calibration efficiency.

The impact of the co-location period on calibration accuracy has also been investigated. The duration required for a stable calibration can vary by sensor type and is influenced by environmental variability and cross-sensitivities. Levy Zamora et al. \cite{levy} found that while improvements in model performance diminished for co-location periods longer than six weeks, the representativeness of environmental conditions during calibration was more critical than the absolute duration, indicating that a strategically chosen period that captures a diverse range of ambient operation conditions (say w.r.t. weather, pollution levels, etc) can be sufficient for achieving robust calibration, even if of a shorter duration.

%% file: setting.tex
\section{Data Collection and Problem Setting}
\label{sec:setting}

\mypar{Deployments} Data was gathered at four deployment sites in the city of Lucknow namely \textbf{B}abasaheb Bhimrao Ambedkar University (26.77◦N, 80.92◦E), \textbf{C}entral Institute of Medicinal and Aromatic Plants (26.89◦N, 80.98◦E), \textbf{G}omti Nagar (26.86◦N, 81.00◦E) and \textbf{T}alkatora (26.83◦N, 80.89◦E). At each site, data was collected in two seasons, namely \textbf{W}inter (Oct-Dec 2023) and \textbf{S}pring (Feb-Apr 2024).\\

\mypar{Instrumentation} A mobile van, that could transit across sites, was fitted with a variety of instruments. This included an array of 6 LCAQ sensor packages intended to measure ambient levels of several pollutants such as \PM, \CO, \Ox, \NOx, \SOx, apart from meteorological parameters such as temperature and relative humidity (\RH). However, in this work, we focus only on the \CO sensors. The CO-B4 series of \CO sensors manufactured by AlphaSense and temperature sensors manufactured by Bosch were used in the sensor packages. The sensor packages are identified using a unique 4-character code (1FEA, 5092, 5A2D, 5CB6, B5FF, D3EB). The van was also fitted with regulatory-grade instruments (RGI) \cite{bousiotis2021assessing} such as an Environmental Beta Attenuation Mass (E-BAM) monitor to measure \PM concentration, and several gas analyzers to measure the concentrations of gases such as \CO, \Ox, \NOx, \SOx \cite{oilandgas_bseries}. However, for this work, we require reference readings only from the \CO gas analyzer. Standard QA/QC steps were taken to calibrate the gas analyzers using zero air and reference concentrations.\\

\mypar{Final Datasets} The sites and seasons were uniquely identified using the first letter of their names i.e. site \textbf{B}, \textbf{C}, \textbf{G}, \textbf{T} and season \textbf{W}, \textbf{S}. This resulted in a total of 8 datasets that are uniquely identified by the site and season letters, separated by a hyphen. For example, data from the Talkatora site collected during the winter season is identified as \textbf{T-W}. For each dataset, timestamped values of measurements from all LCAQ and meteorological sensors from the 6 sensor packages was available alongside measurements offered by the RGI. The LCAQ sensors offered measurements at 1-minute intervals whereas the RGI offered data at 15-minute intervals. To resolve this, LCAQ data was averaged at 15-minute intervals.\\

\mypar{Problem Statement} The sensing technology used in LCAQ CO gas sensors is electrochemical in nature \cite{datasheet}. For each timestamp $t$, the reference-grade gas analyzer provides a single reading say $y_t$ for the ambient \CO level. On the other hand, the LCAQ sensor provides at each timestamp, two operating potential (OP) values, say $\OP_{1t}, \OP_{2t}$ reported by the working and auxiliary electrodes situated on the sensor that are known to be correlated with the ambient levels of \CO. Ambient temperature values $T_t$ are also required to be measured since the sensitivity of sensor is known to be temperature dependent. Given this, the goal of LCAQ sensor calibration can be cast as a regression problem, that of estimating a function $f$ of the operating potentials and ambient temperature values that well-approximates the readings offered by the RGI gas analyzer i.e. $y_t \approx f(\OP^1_t, \OP^2_t, T_t)$.

%% file: method.tex
\section{\alg: outlier \textsf{RE}sistant \textsf{S}emi \textsf{P}arametr\textsf{I}c \textsf{RE}gression}
\label{sec:method}
The \alg method is based on two key insights -- firstly, that the calibration model suggested by the sensor manufacturer does not require entirely non-linear models and secondly, being resistant to outliers in training data is beneficial to the calibration process.\\

\mypar{Semi-parametric Regression} For the \CO LCAQ sensor, the auxiliary electrode is not exposed to ambient air whereas the working electrode is exposed to ambient air. Thus, in a perfect scenario, ambient \CO measurements could be obtained from the potential readings simply by subtracting the auxiliary electrode potential from the working electrode potential (to eliminate zero currents) and dividing by the sensitivity value of the sensor. However, as noted by the manufacturer \cite{alphasense-note}, this simple model is frustrated by idiosyncrasies of the electrochemical reactions:
\begin{enumerate}
	\item The effective surface areas of the electrodes are non-identical requiring multipliers to be introduced.
	\item The surface areas are know to vary with temperature due to expansion and contraction in the membrane pores, making these multipliers temperature dependent.
	\item The sensitivity of the sensor varies with temperature but the variation is not described in a simple closed-form expression and merely known via empirical graphs \cite{datasheet}
\end{enumerate}
Although both linear \cite{hoerl1970}, non-linear \cite{scholkopf1998,quinlan1986,friedman2001,rumelhart1986,cover1967}, and even non-parametric models have been proposed for calibration, it notable that the calibrations steps suggested by the manufacturer requires the model to be only partially non-linear. \alg takes advantage of this -- let $x_{1t}, x_{2t}$ denote the two operating potentials recorded by the LCAQ \CO sensor. Then \alg models the ambient \CO levels by treating the sensitivity parameters as \emph{nuisance variables} and modeling them non-parametrically
\[
y_t \approx w_1(T_t) \cdot x_{1t} + w_2(T_t) \cdot x_{2t} + b(T_t)
\]
where the weights $w_1(T_t), w_2(T_t)$ capture the temperature dependence of both the multiplier and sensitivity and $b(T_t)$ is a temperature dependent bias term. As the nature of dependence of the multiplier and sensitivity on ambient temperature is unknown, \alg models both {nuisance parameters} as non-parametric functions of temperature which serves as an \emph{auxiliary} variable in this setting, thus yielding a semi-parametric model.\\

\mypar{Training Algorithm} To avoid notational clutter, we first describe the algorithm for a simplified model where the independent variables are univariate and the auxiliary variable is denoted as $z$ i.e. $y \approx w(z) \cdot x + b(z)$. The case of multivariate independent variables is similar and discussed thereafter. Let there be $N$ timestamps. Let $\vy \in \bR^N$ denote the vector of ground truth values at various timestamps (\CO levels as measured by the RGI gas analyzer). Let $X \in \bR^{N \times N}$ denote a diagonal matrix with the values of the independent variable $x$ at various timestamps in the diagonal. Let $\vw, \vb \in \bR^N$ be vectors denoting the weights and bias values at various timestamps that depend on the value of the auxiliary variable $z$ at that respective timestamp. To implement a non-parametric model for the weights and biases, we identify a Mercer kernel $\cK$ with corresponding RKHS $\cH$ and feature map $\phi: \bR \rightarrow \cH$. Let $Z: \cH \mapsto \bR^N$ denote an operator composed as $Z \deff \bs{\phi(z_1),\phi(z_2),\ldots,\phi(z_N)}^\top$ where $\phi(z_t) \in \cH$ is the feature map of the auxiliary variable at timestamp $t$. Given this, the weights and biases can be modeled as $\vw = Z\vp, \vb = Z\vq$ where $\vp,\vq \in \cH$. The predictions given by this model are of the form
\[
\hvy = X\vw + \vb = XZ\vp + Z\vq
\]
Using the least squares loss along with Hilbertian regularization (where $\norm\cdot_{\cH}$ denotes the Hilbertian norm in the RKHS $\cH$) yields the following Tikhonov-style optimization problem
\[
\min_{\vp,\vq\in\cH}\ \frac\lambda2\cdot(\norm\vp_{\cH}^2 + \norm\vq_{\cH}^2) + \norm{\vy - XZ\vp - Z\vq}_2^2
\]
Given the intractability of solving the primal directly, we introduce an auxiliary variable representing the residual $\vr$, a consequent constraint, namely $\vr = \vy - XZ\vp - Zq$, and a corresponding Lagrangian dual variable $\vbeta \in \bR^N$. Following standard steps, the following Lagrangian dual problem is obtained.
\[
\min_{\vbeta \in \bR^N}\ \vbeta^\top (G + H) \vbeta + \lambda\cdot\norm\vbeta_2^2 - 2\lambda\cdot\vbeta^\top\vy,
\]
where $G \deff ZZ^\top \in \bR^{N \times N}$ is the Gram matrix over the auxiliary variables and $H \deff XGX \in \bR^{N \times N}$. Once the model has been learnt, making predictions on novel points is straightforward.\\

\mypar{Inference} Given a novel point $(x, z) \in \bR \times \bR$ with independent variable $x$ and auxiliary variable $z$, this model can make a prediction as
\[
\hat y = w(z)\cdot x + b(z)
\]
To obtain the weight and bias values $w(z), b(z)$, we note that the standard KKT conditions ensure Lagrangian stationarity with respect to the primal variables that gives us the following relations
\begin{align*}
    \vp &= \frac{Z^\top X^\top\vbeta}\lambda\\
    \vq &= \frac{Z^\top\vbeta}\lambda
\end{align*}
Thus, for the novel point, we get
\begin{align*}
    w(z) &= \phi(z)^\top\vp = \frac{\ip{\vk}{X\vbeta}}\lambda\\
    b(z) &= \phi(z)^\top\vq = \frac{\ip{\vk}{\vbeta}}\lambda,
\end{align*}
where $\vk = [\cK(z,z_1), \cK(z,z_2), \ldots, \cK(z,z_N)]^\top = \phi(z)^\top Z^\top \in \bR^N$ is the vector of kernel values of the novel auxiliary variable with the train auxiliary variables. Thus, inference can be performed by computing
\[
\hat y = \br{\frac{\ip{\vk}{X\vbeta}}\lambda}\cdot x + \br{\frac{\ip{\vk}{\vbeta}}\lambda}
\]

\mypar{Extensions to Multivariate Inputs} To handle calibration of \CO sensors, we need to handle cases where the independent variable is multivariate, i.e., say each time produces a tuple of the form $(x_1, x_2, z)$ where $z$ is the auxiliary variable and $(x_1, x_2)$ is the multivariate independent variable. This is because the sensor records two potentials $\OP_1, \OP_2$ at each timestamp. In this case, the output is modeled as 
\[
y \approx w_1(z) \cdot x_1 + w_2(z) \cdot x_2 + b(z)
\]
In vectorized notation, the predictions of the model would be of the form
\[
\hvy = X_1\vw_1 + X_2\vw_2 + \vb = X_1Z\vp_1 + X_2Z\vp_2 + Z\vq,
\]
where $\vw_1 = Z\vp_1, \vw_2 = Z\vp_2, \vb = Z\vq$ where $\vp_1,\vp_2,\vq \in \cH$ and $X_1, X_2 \in \bR^{N \times N}$ are two diagonal matrices containing the two independent variables $x_1, x_2$ in their respective diagonals. The primal stationarity KKT conditions guarantee $\vp_i = \frac{Z^\top X_i^\top\vbeta}\lambda$ for $i = 1, 2$, resulting in a dual of the form
\[
\min_{\vbeta \in \bR^N}\ \vbeta^\top (G + H_1 + H_2) \vbeta + \lambda\cdot\norm\vbeta_2^2 - 2\lambda\cdot\vbeta^\top\vy,
\]
where $H_i \deff X_iGX_i \in \bR^{N \times N}$ for $i = 1, 2$. Thus, the semi-parametric procedure outlined above extends to the multivariate case simply by replacing $H \deff X_1 G X_1 + X_2 G X_2$. The inference step is similarly modified as follows:
\[
\hat y = \br{\frac{\ip{\vk}{X_1\vbeta}}\lambda}\cdot x_1 + \br{\frac{\ip{\vk}{X_2\vbeta}}\lambda}\cdot x_2 + \br{\frac{\ip{\vk}{\vbeta}}\lambda}
\]
It is easy to see that the above procedure can readily handle any number of independent variables i.e., cases where each timestamp offers an auxiliary variable $z$ as well as $x_1, x_2, x_3, \ldots, x_d$ where $d \geq 2$.\\

\mypar{Model Compression} Once a model $\vbeta \in \bR^N$ has been learnt, \alg can compress the model by sparsifying it. Although various model compression techniques for kernel methods have been proposed in literature, \alg adopts a straightforward hard-thresholding-based technique. First it chooses the top $\hat N$ coordinates in $\vbeta$ for some value of $\hat N < N$ and sets the rest to 0, thus retaining only the most influential support vectors. The coordinates retained are fined tuned by solving a simple least squares problem yielding a sparse model with model size of $\hat N \ll \bigO N$. In experiments, this compression does not adversely impact performance (see Figure~\ref{fig:compression_r2_drop}.

\begin{figure}[t]
    \centering
        \begin{subfigure}[t]{0.49\textwidth}
        \caption{\underline{\textbf{\alg Training}}}
        \begin{algorithmic}[1]
        \STATE \textbf{Input:} Independent vars $X_1, X_2 \in \bR^{N \times N}$, Gram matrix over auxiliary vars $G \in \bR^{N \times N}$, target $\vy  \in \bR^N$, regularization $\lambda$, corruption fraction $\alpha \in [0,1]$, correction rate $\eta \in [0,1]$
        \STATE \textbf{Output:} Predictor parameters $\vm, \vn, \vo \in \bR^N$
        \STATE $\vc \leftarrow \vzero$ \COMMENT{Initialize null corruption}
        \WHILE{not converged}
            \STATE $\{\vm,\vn,\vo\} \leftarrow \text{\textsf{SPR}}(X_1, X_2, G, \vy - \eta\cdot\vc, \lambda)$
            \STATE $\hat\vy \leftarrow X_1G\vm + X_2G\vn + G\vo$
			\STATE $\vc \leftarrow \text{\textsf{HT}}(\vy - \hat\vy, \alpha\cdot N)$
        \ENDWHILE
        \STATE \textbf{return} $\vm, \vn, \vo$
        \end{algorithmic}
        \vspace*{1ex}
        \caption{\underline{\textbf{\alg Inference}}}
        \begin{algorithmic}[1]
        \STATE \textbf{Input:} Test indep. vars $x_1, x_2 \in \bR$, aux kernel values $\vk \in \bR^N$, learnt params $\vm, \vn, \vo \in \bR^N$
        \STATE \textbf{Output:} Prediction $\hat y$
        \STATE \textbf{return} $\hat y \leftarrow\vk^\top\vm\cdot x_1 + \vk^\top\vn\cdot x_2 + \vk^\top\vo$
        \end{algorithmic}
    \end{subfigure}
    \hfill
    \begin{subfigure}[t]{0.49\textwidth}
        \caption{\underline{\textbf{Semi-parametric Regression (\textsf{SPR})}}}
        \begin{algorithmic}[1]
        \STATE \textbf{Input:} Independent vars $X_1, X_2 \in \bR^{N \times N}$, Gram matrix over auxiliary vars $G \in \bR^{N \times N}$, target $\vy  \in \bR^N$, regularization hyperparam. $\lambda$
        \STATE \textbf{Output:} Predictor parameters $\vm, \vn, \vo \in \bR^N$
        \STATE Solve dual objective to obtain $\vbeta$
        \STATE \textbf{return} $\vm \deff \frac{X_1\vbeta}\lambda, \vn \deff \frac{X_2\vbeta}\lambda, \vo \deff \frac\vbeta\lambda$
        \end{algorithmic}
        \vspace*{6ex}
        \caption{\underline{\textbf{Hard Thresholding (\textsf{HT})}}}
        \begin{algorithmic}[1]
        \STATE \textbf{Input:} Vector $\vr \in \bR^N$, sparsity $k \in [N]$.
        \STATE \textbf{Output:} Sparsified vector $\vz$
        \STATE Let $j_1, j_2, \ldots, j_N \in [N]$ be the coordinates of $\vr$ in decreasing order of magnitude i.e. $\abs{r_{j_1}} \geq \abs{r_{j_2}} \geq \ldots \geq \abs{r_{j_N}}$
        \STATE Initialize $\vz \leftarrow \vzero$
        \STATE Assign $z_{j_i} = r_{j_i}$ for all $i \in [k]$
        \STATE \textbf{return} $\vz$
        \end{algorithmic}
    \end{subfigure}
    \caption{\alg pseudocode describing the base training and inference procedures and robust learning}
    \label{fig:alg}
\end{figure}

\subsection{Outlier-resistant Regression}
Real-life deployments present several sources of outliers in data. Intermittent, short-lived spikes in \CO levels present a challenge for calibration as LCAQ sensors may have slower response times as well smaller dynamic ranges given their modest construction. Given this, outlier-resistance becomes a valuable trait. Although robust regression methods with provable breakdown points have been extensively studied for linear models \cite{maronna2011, DalalyanChen2012, ChenCaramanisMannor2013, McWilliamsKrummenacherLucicBuhmann2014, BhatiaJainKar2015, DiakonikolasKongStewart2019, MukhotyGJK19}, extensions to robust non-parametric regression are far less studied. We are not aware of any works that explicitly offer robustness guarantees for semi-parametric models. There are some works \cite{FanHuTruong1994, DuWangBalakrishnanRavikumarSingh2018, CizekSadikoglu2020, Mukhoty2021} that address robust non-parametric regression but neither address semi-parametric regression explicitly and most do not offer explicit breakdown points.

\alg develops its robust semi-parametric regression (SPR) method by adapting the APIS method originally proposed in \cite{Mukhoty2021} non-parametric regression. The APIS method is a meta-algorithm that, when adapted to the case of semi-parametric regression, results in the algorithm outlined in Figure~\ref{fig:alg}. Suppose it is anticipated that an $\alpha$-fraction of training points are outliers. Let \textsf{SPR} denote an execution of the semi-parametric regression procedure outlined in Figure~\ref{fig:alg}(c) that yields prediction coefficients that can be used to make predictions on data points using the procedure outlined in Figure~\ref{fig:alg}(b). Also, for any vector $\vr \in \bR^N$ and any $k < N$, let the \emph{hard thresholding} operation $\HT(\vr, k)$ be defined as yielding an $k$-sparse vector obtained by retaining the $k$ coordinates of $\vv$ with highest magnitude (irrespective of sign) as is and setting the other $N-k$ coordinates to zero, as outlined in Figure~\ref{fig:alg}(d). \alg performs outlier-resistant semi-parametric regression by initializing a \emph{corruption estimate} vector $\vc$ to zero and then alternating between learning the prediction parameters using the \emph{corrected} targets, and re-estimating the correction vector, as outlined in Figure~\ref{fig:alg}(a). The \emph{outlier correction rate} hyperparameter $\eta$ controls how aggressively are the estimated sparse outliers shaved off from the ground truth before proceeding and the value of $\eta$ is tuned as a hyperparameter (see Section~\ref{sec:exps}).\\

\begin{figure*}[t]
    \centering
    \begin{subfigure}[b]{0.32\textwidth}
        \centering
        \includegraphics[width=\textwidth]{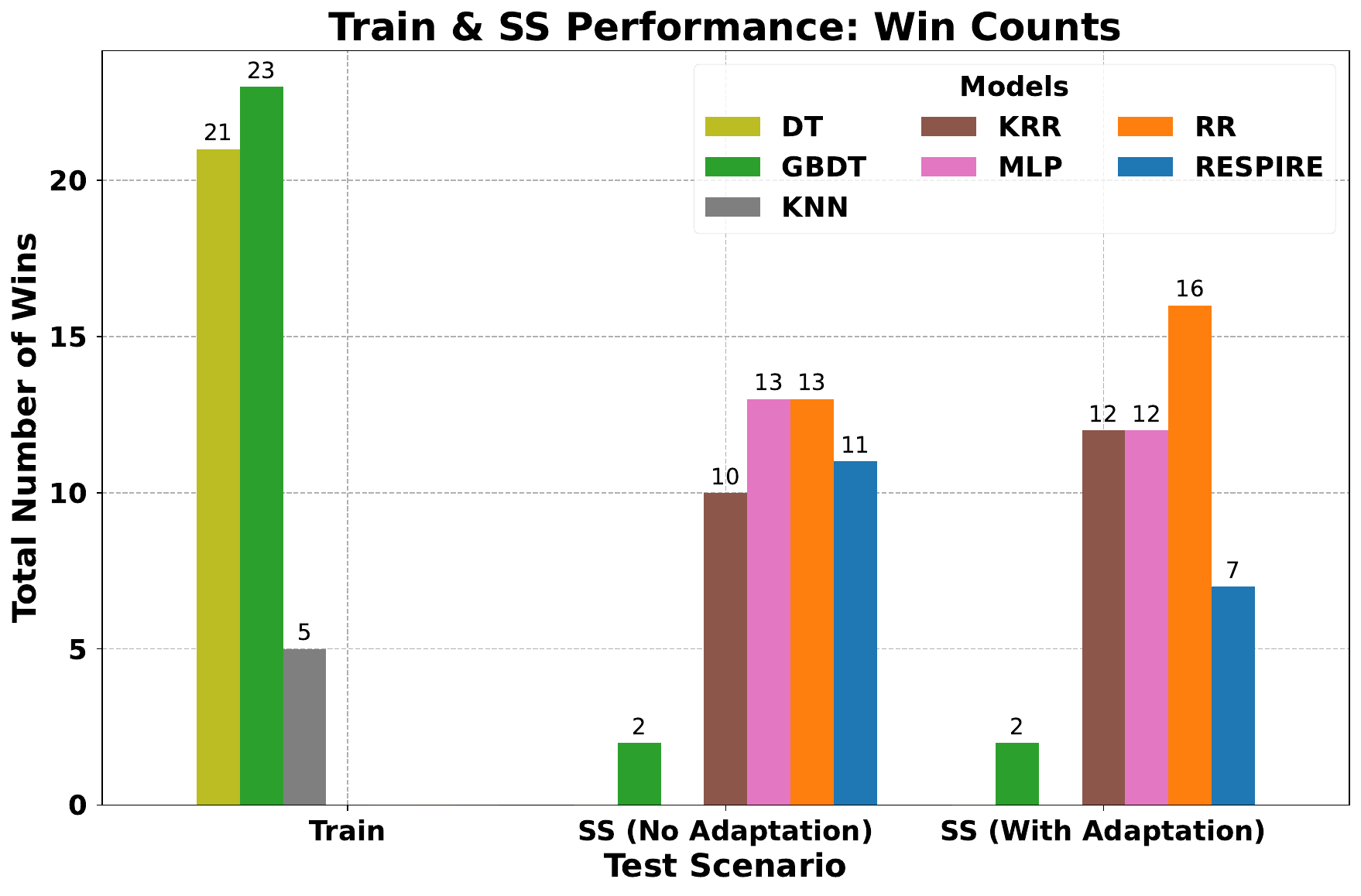}
        \caption{Win counts for various methods on train data as well as non-transfer (SS) scenarios, both with and without adaptation.}
        \label{fig:sub2_bg}
    \end{subfigure}
    \hfill
    \begin{subfigure}[b]{0.32\textwidth}
        \centering
        \includegraphics[width=\textwidth]{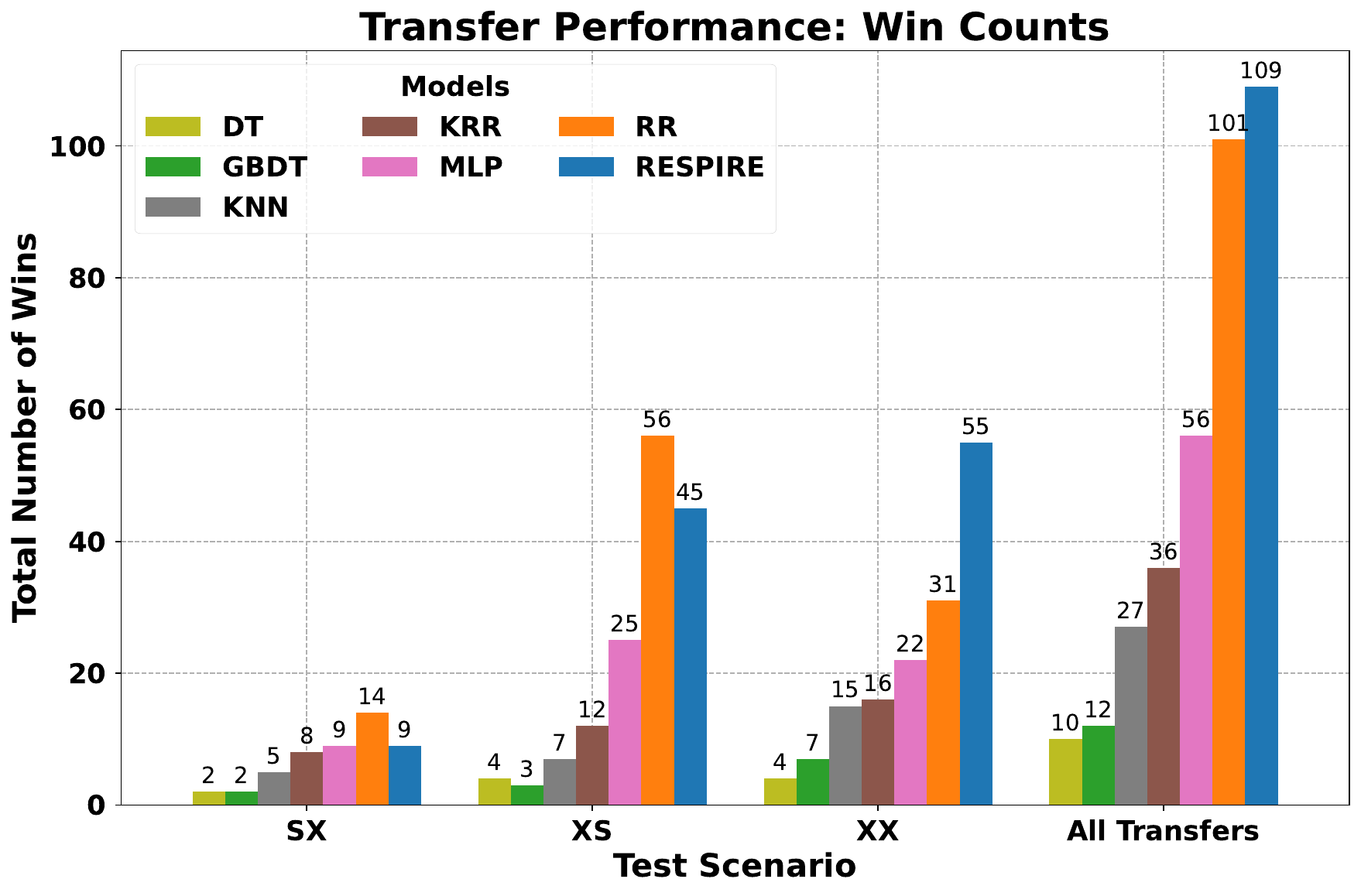}
        \caption{Win counts on transfer (SX, XS, XX) cases and total win counts when no adapter was offered to any method.}
        \label{fig:transfer_no_adapt}
    \end{subfigure}
    \hfill
    \begin{subfigure}[b]{0.32\textwidth}
        \centering
        \includegraphics[width=\textwidth]{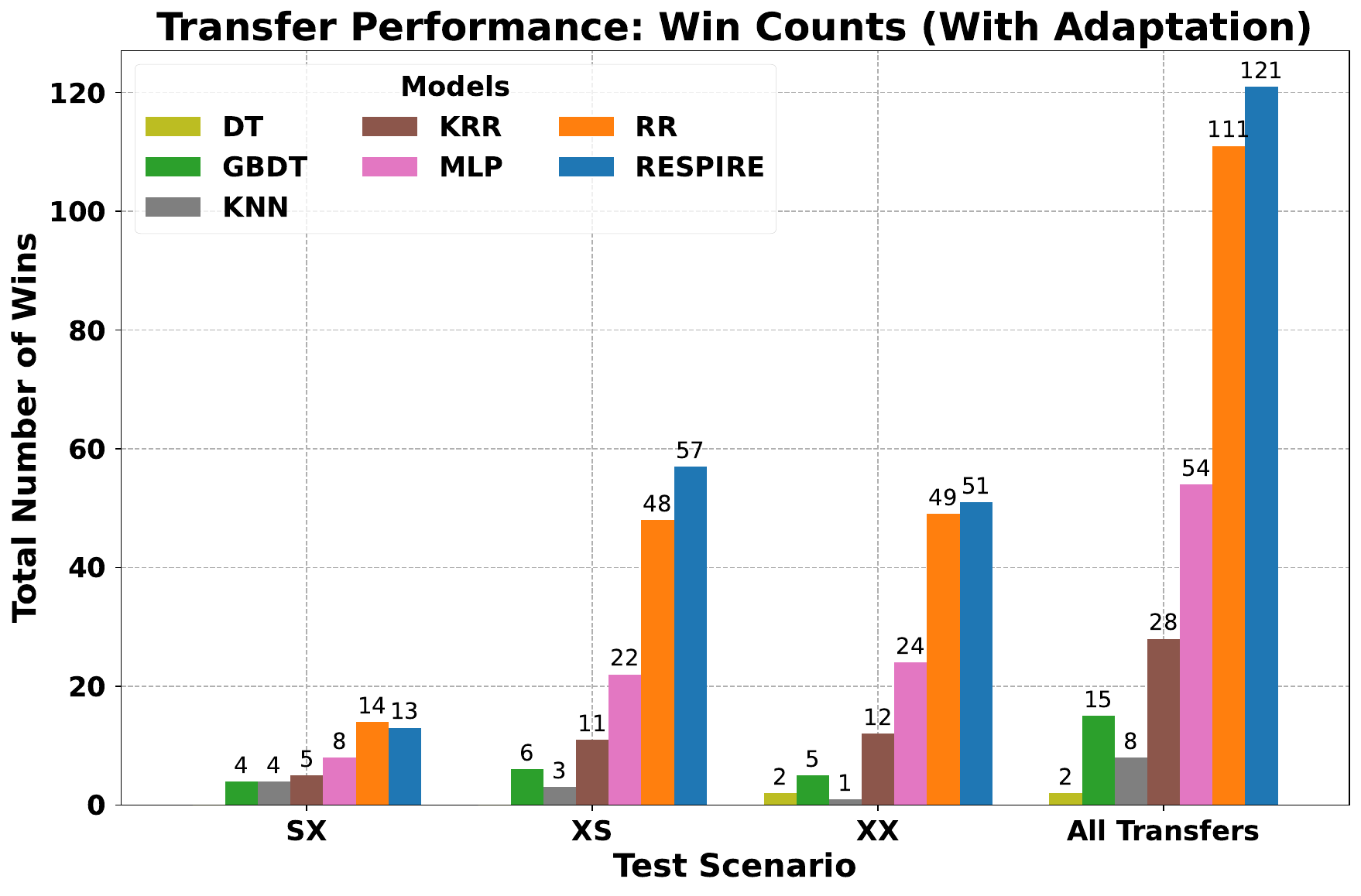}
        \caption{Win counts on transfer (SS, XS, XX) cases and total win counts when a 1D adapter was offered to all methods.}
        \label{fig:transfer_adapt}
    \end{subfigure}
    \caption{Leader boards for train, non-transfer (SS) and transfer (SX, XS, XX) experiments. The height of the bars indicate how many times a method delivered the (tied) best performance across all 6 sensors, on all sites during training, SS, SX, XS and XX experiments. The baseline methods DT, GBDT, KNN dominate train performance but perform poorly even in the SS transfer case where MLP, KRR and linear (RR) dominate. However, when considering non-trivial transfers such as SX, XS and XX, \alg is the clear winner whether adaptation is offered or not. MLP and linear (RR) are the next best methods.}
    \label{fig:row1}
\end{figure*}

\begin{figure*}[t]
    \centering
    \begin{subfigure}[b]{0.9\textwidth}
        \centering
        \includegraphics[width=\textwidth]{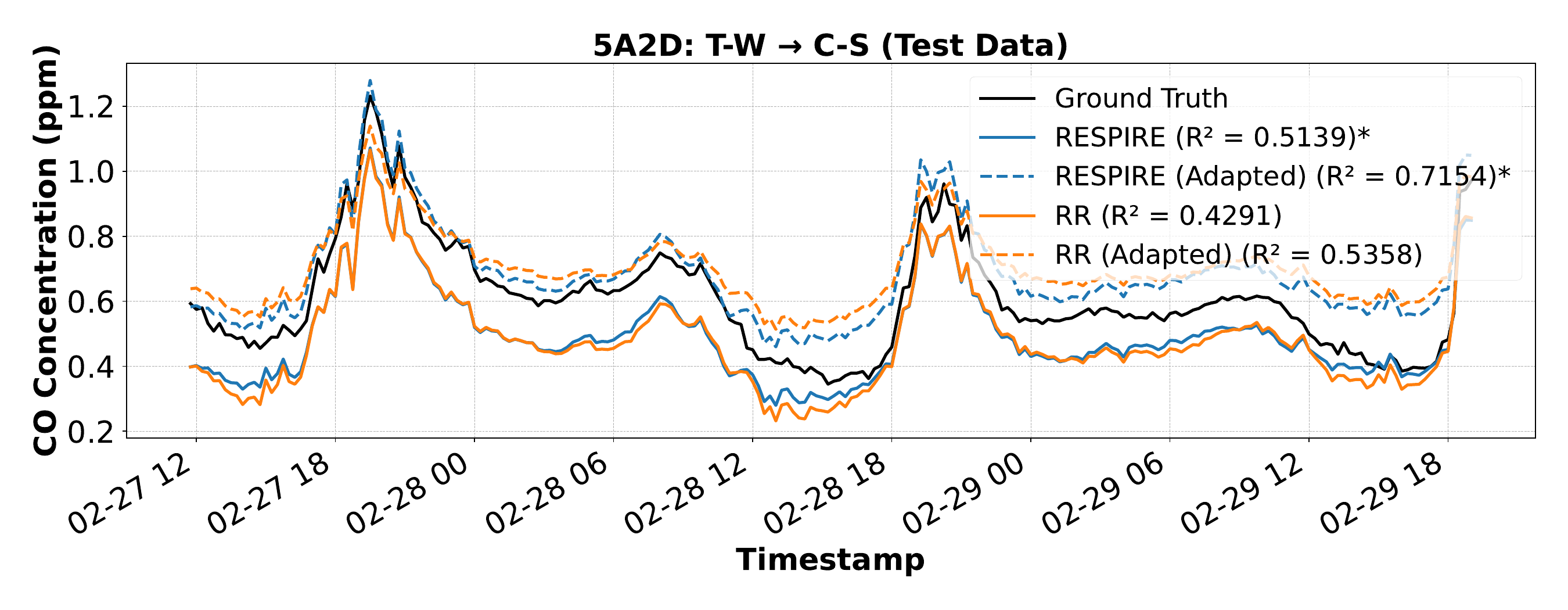}
        \caption{An XX transfer scenario with training done at site T on winter data and testing at site C on spring data. Notice how the model predictions closely track the ground truth but offer poor \metric scores due to bias errors. Once the adapter is applied, excellent \metric scores are achieved. \alg wins on this transfer with linear (RR) being the next best method.}
        \label{fig:sub1_ts}
    \end{subfigure}
    \hfill
    \begin{subfigure}[b]{0.9\textwidth}
        \centering
        \includegraphics[width=\textwidth]{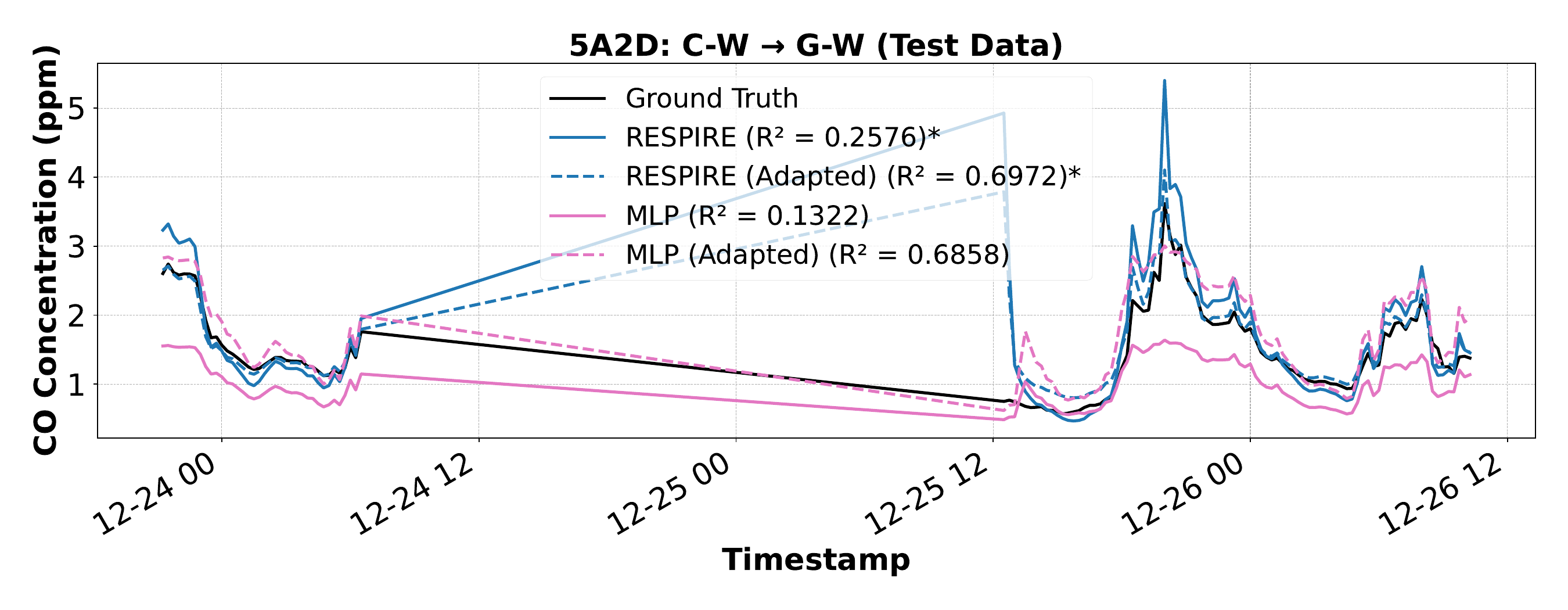}
        \caption{An XS transfer scenario during winter with training done on site C data and testing on site G data. Notice how the model predictions closely track the ground truth but offer poor \metric scores due to a bias error. Once the adapter is applied, excellent \metric is achieved. \alg wins on this transfer with MLP being the next best method.}
        \label{fig:sub2_ts}
    \end{subfigure}
    \caption{Two transfer scenarios demonstrating possible zero-shifts in RGI data being a contributor to poor \metric performance. A simple 1D adapter greatly improves performance in all cases for all methods.}
    \label{fig:row2}
\end{figure*}

\mypar{Breakdown Point Analysis}
Although the APIS meta algorithm  does extend to semi-parametric models, the breakdown point analysis presented in \cite{Mukhoty2021} does not. The breakdown analysis for \alg is presented below under the following assumptions.
\begin{enumerate}
    \item \textbf{Adversary Model}: the noise model considered in the calculations is the so-called \emph{fully adversarial} model wherein an adversary can take choose to corrupt an arbitrary $\alpha$ fraction of the training targets, i.e., corrupt $\alpha\cdot N$ coordinates of the training label vector $\vy \in \bR^N$, in an arbitrary manner. To decide which coordinates to corrupt and by how much, the adversary is allowed full knowledge of the independent and auxiliary covariates i.e. $X_1, X_2, G$, as well as the true (uncorrupted) label vector $\vy$. It is notable that the covariates are assumed to suffer no adversarial corruption.
    \item \textbf{Full Correction}: the analysis is presented for the setting $\eta = 1$ i.e. the estimated (sparse) outliers are completely eliminated during the correction step. This is required because the analysis assumes adversarial corruption. In practice, the outliers are not adversarial and a milder value of $\eta < 1$ is used after being tuned as a hyperparameter.
    \item \textbf{RBF Kernel}: the calculations are presented for the case of the RBF kernel as (asymptotic) spectral bounds have been widely studied for this kernel.
    \item \textbf{Auxiliary Covariate Distribution} the calculations assume that the auxiliary covariate $z$ is uniformly distributed in the interval $[0,1]$. This is an assumption inherited from spectral calculations for the RBF kernel and mostly serve to make the eigenvalue estimation easier. Real life data can be made to respect the $[0,1]$ interval bound by performing min-max normalization of the auxiliary (temperature) variable. If data does not satisfy the uniformity assumption, this would merely worsen the eigenvalue bounds.
    \item \textbf{Independent Covariate Bound}: the calculations assume that the independent covariates $x_1, x_2$ are bounded in a range $[r,R]$. This is a realistic assumption since for \CO calibration, the independent variables are potentials that take values in a small range (typically in millivolts).
\end{enumerate}

\begin{theorem}
Given independent variables $X_1, X_2$ and Gram matrix $G$ over the auxiliary variables, suppose the (uncorrupted) targets are generated as $\vyo = (X_1GX_1 + X_2GX_2 + G)\vbetao$ where $\vbetao \in \bR^N$ is spanned by the top $s$ eigenvectors of $X_1GX_1 + X_2GX_2 + G$ for $s = \bigO{\frac{\log N}{\log\log N}}$. Suppose the adversary, when presented with $X_1, X_2, G, \vy, \vbetao$, decides to introduce corruptions so that the observed (corrupted) targets are generated as $\vy = \vyo + \vb + \veo$ where and $\vb$ is an arbitrary $k$-sparse corruption signal for $k = \bigO{\sqrt N}$ and $\veo$ denotes benign noise not introduced by the adversary. Then, if the independent covariates $x_1, x_2$ are bounded in an interval, say $[r, R]$ for $r > 0$ and $\frac Rr = \bigO1$, and the auxiliary covariates $z$ are distributed uniformly over the unit interval $[0,1]$ and if the Gram matrix $G$ is generated using the Gaussian kernel with the bandwidth parameter $h \in \bs{\sqrt{\frac{40}{\log N}},1}$, then with probability at least $1 - \bigO{\exp\br{-n^{\frac25}}}$, for any $\epsilon > 0$, the \alg procedure outlined in Figure~\ref{fig:alg} converges to a solution $\hat\vbeta$ satisfying $\norm{\hat\vbeta - \vbetao}_2 < \epsilon + 7\cdot\norm\veo_2$ within $\bigO{\log\frac1\epsilon}$ iterations.
\end{theorem}
\begin{proof}
It is notable that the benign noise $\veo$ can account for various factors such as
\begin{enumerate}
	\item Non-adversarial noise, say Gaussian noise, in the uncorrupted targets
	\item Modelling error, say if the uncorrupted targets are well-explained by the the top $s$ eigenvectors, but require more eigenvectors to be perfectly explained
	\item Non-adversarial covariate noise, say observation noise, in the covariates $x_1, x_2, z$
\end{enumerate}
We first present the proof for the case when there is no benign noise i.e. $\veo = \vzero$ i.e. there is no modelling error, covariate noise or non-adversarial target noise. The claimed result then follows from a direct application of \cite[Lemma 9]{Mukhoty2021} and the fact that orthogonal projections are always contractive.

We first note that the uncorrupted signals can be expressed as $\vyo = F\vbetao$ for $F \deff X_1GX_1 + X_2GX_2 + G$ which represents this as an instance purely non-parametric regression with Gram matrix $F$. Since $X_1, X_2$ are diagonal matrices, $X_iGX_i = G \odot J_i$ where $\odot$ denotes the coordinate-wise multiplication operation and $J_i \deff \vx_i\vx_i^\top$ and $\vx_i \in \bR^N$ is the vector created out of the diagonal entries of $X_i$. Since $J_i$ is clearly a positive semi-definite (PSD) matrix and the set of PSD matrices is closed under pointwise multiplication, $X_iGX_i$ is PSD whenever $G$ is PSD. Since the set of PSD matrices is also closed under addition, $F$ is a PSD matrix whenever $G$ is a PSD matrix. These closure results over PSD matrices are standard but are replicated in Lemma~\ref{lem:psd} for sake of completeness.

\cite[Lemma 5]{Mukhoty2021} establishes that in the noiseless setting i.e. when $\veo = \vzero$, the alternating procedure adopted by \alg offers exact recovery at a linear rate whenever $3\cdot\Lambda^{\text{unif}}_k(F) < \lambda_s(F)$ where $\lambda_s(F)$ is the $s\nth$-largest eigenvalue of $F$ and for any $k > 0$, the quantity $\Lambda^{\text{unif}}_k(F)$ denotes the largest eigenvalue of any principal $k \times k$ sub-matrix of $F$. A principal sub-matrix is one that projects a matrix onto an identical set of rows and columns.

Using Lemma~\ref{lem:horn}, denoting $\Lambda \deff \Lambda^{\text{unif}}_k(G), \lambda \deff \lambda_s(G)$ and using the simple observations $\Lambda^{\text{unif}}_k(X_i) \leq R$ and $\lambda_N(X_i) \geq r$, we get the following results:
\begin{align*}
\Lambda^{\text{unif}}_k(F) &\leq \Lambda\cdot(2R^2 + 1)\\
\lambda_s(F) &\geq \lambda\cdot\max\bc{r^2,1}
\end{align*}
This implies that we require
\[
\frac{\Lambda\cdot(2R^2 + 1)}{\lambda\cdot\max\bc{r^2,1}} \leq \frac13 \Leftrightarrow \frac\Lambda\lambda \leq \frac{\max\bc{r^2,1}}{3(2R^2 + 1)}
\]
Using results from \cite[Theorem 2]{MinhNiyogiYao2006} and calculations from \cite[Section C.3]{Mukhoty2021}, the above is assured for the claimed ranges of $s, k$ whenever $\frac Rr = \bigO1$.

It is notable that in the noiseless case $\veo = \vzero$, \alg assures exact recovery, and a consistent estimate, at a linear rate of convergence. For noisy, ill-modeled case, the modelling error or stochastic noise is inherited into the estimate with a small constant multiplier.
\end{proof}

\begin{lemma}
\label{lem:horn}
For any square, symmetric positive semi-definite matrix $A \in \bR^{N \times N}$, for any $s \in [N]$, let $\lambda_s(A)$ be the $s\nth$-largest eigenvalue of $A$ and for any $k \in [N]$, let $\Lambda^{\text{unif}}_k(A)$ denote the largest eigenvalue of any principal $k \times k$ sub-matrix of $F$. Furthermore, let $A, B \in \bR^{N \times N}$ denote two square, symmetric PSD matrices and $D \in \bR^{N \times N}$ denote an invertible diagonal matrix. Then
\begin{enumerate}
	\item $\lambda_s(A + B) \geq \max\bc{\lambda_s(A), \lambda_s(B)}$
	\item $\Lambda^{\text{unif}}_k(A + B) \leq \Lambda^{\text{unif}}_k(A) + \Lambda^{\text{unif}}_k(B)$
	\item $\lambda_s(DAD) \geq \lambda_s(A)\cdot\lambda_N(D)^2$
	\item $\Lambda^{\text{unif}}_k(DAD) \leq \Lambda^{\text{unif}}_k(A)\cdot(\Lambda^{\text{unif}}_k(D))^2$
\end{enumerate}
\end{lemma}
\begin{proof}
We prove the parts separately.
\begin{enumerate}
	\item Using the Horn's inequalities \cite{KnutsonTao2001Honeycombs}, we get $\lambda_s(A + B) \geq \max_{i + j = n - s}\ \br{\lambda_{n-i}(A) + \lambda_{n-j}(B)} \geq \max\bc{\lambda_s(A), \lambda_s(B)}$
	\item Let $S \subset [N], \abs S = k$ be the subset of rows and columns over which the values is achieved i.e. $\Lambda^{\text{unif}}_k(A + B) = \lambda_1\br{(A + B)_S} = \lambda_1(A_S + B_S)$. Using the Horn's inequalities gives us $\lambda_1(A_S + B_S) \leq \lambda_1(A_S) + \lambda_1(B_S) \leq \Lambda^{\text{unif}}_k(A) + \Lambda^{\text{unif}}_k(B)$ where the last step follows from the definition of $\Lambda^{\text{unif}}_k$.
	\item The matrices $DAD$ and $ADD$ share all eigenvalues since they are similar as $D$ is assumed to be invertible. Using extensions of the Horn's inequalities to products of matrices \cite{Klyachko}, we get $\lambda_s(ADD) \geq \lambda_s(A)\cdot\lambda_N(DD) = \lambda_s(A)\cdot(\lambda_N(D))^2$ where the last step follows as $D$ is a diagonal matrix.
	\item Let $S \subset [N], \abs S = k$ be the subset of rows and columns over which the values is achieved i.e. $\Lambda^{\text{unif}}_k(DAD) = \lambda_1\br{(DAD)_S}\lambda_1\br{D_SA_SD_S} = \lambda_1\br{A_SD_SD_S}$. This holds since $D$ is a diagonal matrix and the arguments from part 3. Using the product Horn's inequalities gives us $\lambda_1\br{A_SD_SD_S} \leq \lambda_1\br{A_S}\cdot\lambda_1\br{D_SD_S} = \lambda_1\br{A_S}\cdot(\lambda_1\br{D_S})^2 \leq \Lambda^{\text{unif}}_k(A)\cdot(\Lambda^{\text{unif}}_k(D))^2$ where the last step follows from the definition of $\Lambda^{\text{unif}}_k$. \qedhere
\end{enumerate}
\end{proof}

\begin{lemma}
\label{lem:psd}
For any two square, symmetric positive semi-definite matrices $A, B \in \bR^{N \times N}$, the matrices $A + B$ and $A \odot B$ are also positive semi-definite where $\odot$ denotes coordinate-wise multiplication.
\end{lemma}
\begin{proof}
We will use two different strategies to show the two parts. To show the first part, note that for any $\vx \in \bR^N$, we have $\vx^\top(A + B)\vx = \vx^\top A\vx + \vx^\top B\vx \geq 0$ since $A, B$ are individually PSD. This establishes that $A + B$ must be PSD too. For the second part, we note that PSD-ness entails that there must exist matrices $U, V \in \bR^{N \times N}$, not necessarily symmetric or PSD themselves, such that $A = UU^\top, B = VV^\top$. To show that $A \odot B$ is PSD, we will construct a matrix $W$ such that $A \odot B = WW^\top$. Let $\vu_i, \vv_i \in \bR^N$ denote the $i\nth$ rows of the matrices $U, V$ respectively. Then construct $\vw_i \deff \vu_i \otimes \vv_i \in \bR^{N^2}$ where $\otimes$ denotes the Kronecker product. For any $i, j \in [N]$, we then have
\begin{align*}
\vw_i^\top\vw_j &= (\vu_i \otimes \vv_i)^\top(\vu_i \otimes \vv_i) = \trace\br{\br{\vu_i\vv_i^\top}^\top\br{\vu_i\vv_i^\top}}\\
&= \trace\br{\vv_i\vu_i^\top\vu_i\vv_i^\top} = \br{\vu_i^\top\vu_i}\br{\vv_i^\top\vv_i}
\end{align*}
The above shows that for any $i, j \in [N]$, the $(i,j)\nth$ entry of $WW^\to$ is the product of the $(i,j)\nth$ entries of the matrices $A$ and $B$. This finishes the proof.
\end{proof}

%% file: exps.tex
\begin{table}[t]
	\centering
	\caption{Number of data points per dataset for each sensor}
	\label{tab:stats}
	\begin{tabular}{@{}ccccccc@{}}
		\toprule
		\textbf{Dataset} & \multicolumn{6}{c}{\textbf{Sensor ID}} \\
		\cmidrule(l){2-7}
		\textbf{Site-Season} & {5A2D} & {5CB6} & {1FEA} & {5092} & {B5FF} & {D3EB} \\
		\midrule
		C-W & 744  & 729  & 439  & 787  & 907  & 907  \\
		C-S & 1102 & 564  & 283  & 714  & 1111 & 974  \\
		B-W & 783  & 784  & 784  & 784  & 784  & 784  \\
		B-S & 1068 & 1070 & 1069 & 1070 & 1070 & 1070 \\
		T-W & 2430 & 2431 & 519  & 2431 & 2431 & 2431 \\
		T-S & 1315 & 1318 & 783  & 1318 & 1318 & 1304 \\
		G-W & 629  & 973  & 972  & 973  & 973  & 973  \\
		G-S & 1122 & 1123 & 1025 & 1123 & 1123 & 1123 \\
		\bottomrule
	\end{tabular}
\end{table}

\begin{figure*}[t]
    \centering
    \begin{minipage}{\textwidth}
        \centering
        \begin{minipage}[b]{0.32\textwidth}
            \centering
            \includegraphics[width=\linewidth]{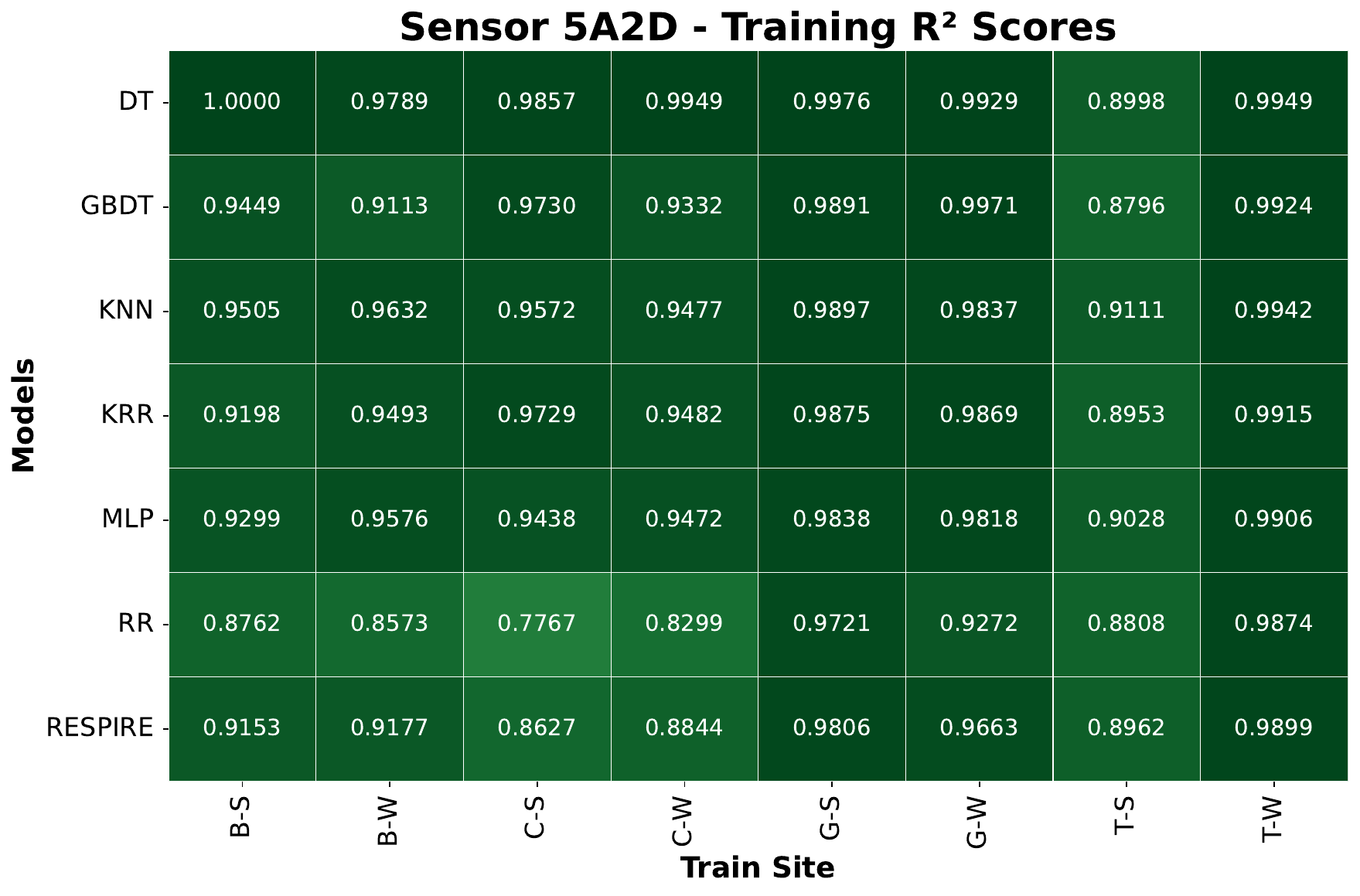}
            \captionof{figure}{All methods offer excellent performance during training.}
            \label{fig:training_performance}
        \end{minipage}
        \hfill
        \begin{minipage}[b]{0.32\textwidth}
            \centering
            \includegraphics[width=\linewidth]{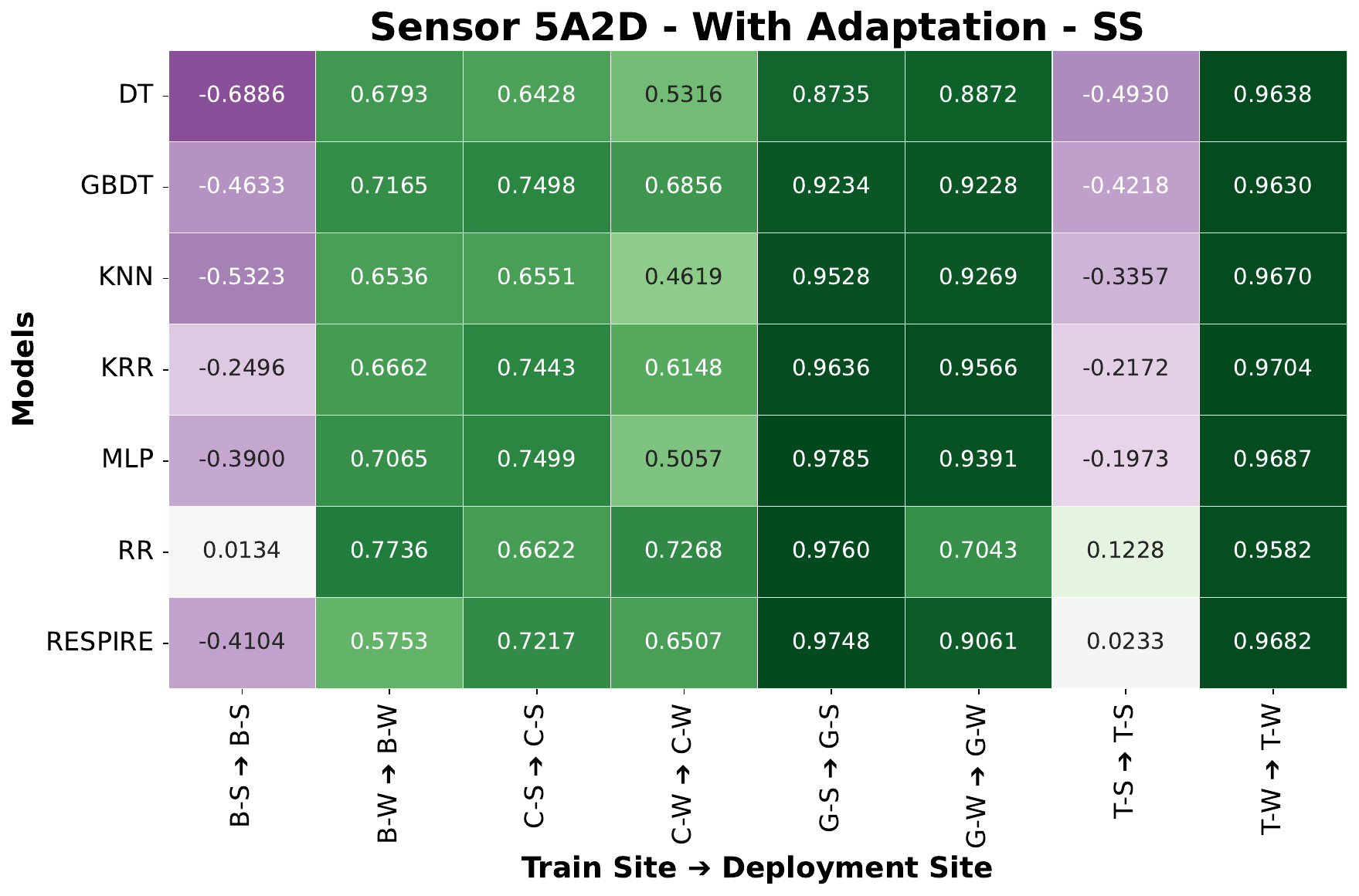}
            \captionof{figure}{DT, GBDT, KNN show steep drops even with non-transfer SS testing.}
            \label{fig:heatmap_ss}
        \end{minipage}
        \hfill
        \begin{minipage}[b]{0.32\textwidth}
            \centering
            \includegraphics[width=\linewidth]{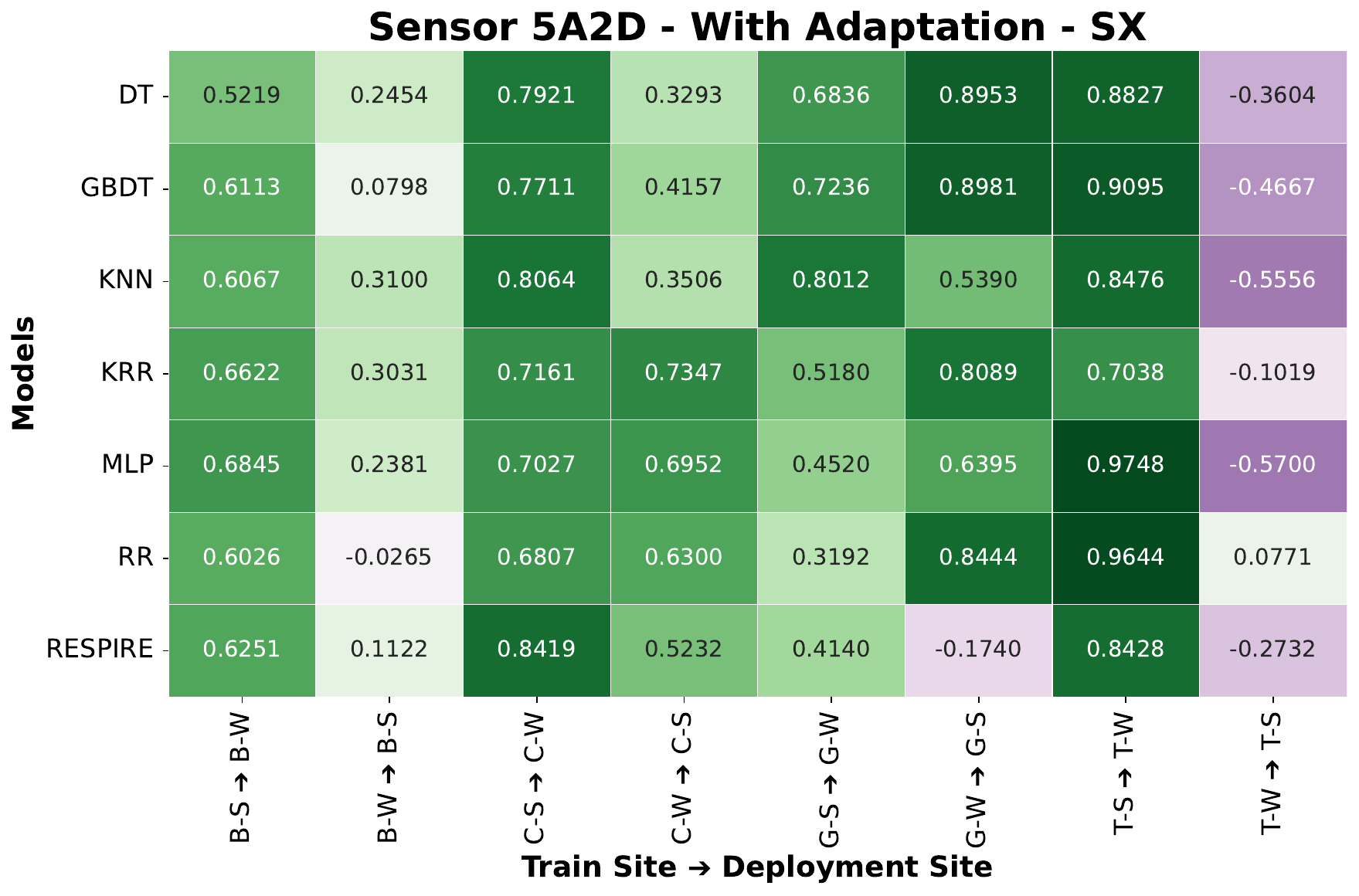}
            \captionof{figure}{\alg offers comparable performance in all SX transfer cases.}
            \label{fig:heatmap_sx}
        \end{minipage}
    \end{minipage}

    \vspace{1em}

    \newsavebox{\spanningblock}
    \sbox{\spanningblock}{%
        \begin{minipage}{0.9\textwidth}
            \centering
            \includegraphics[width=\linewidth]{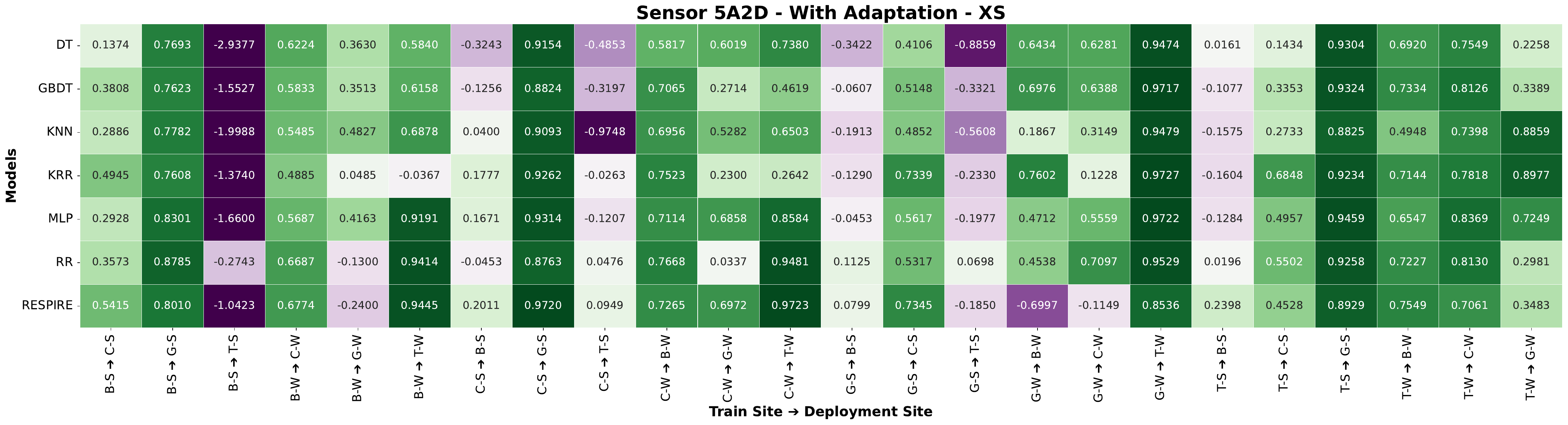}
            \vspace{1em}
            \includegraphics[width=\linewidth]{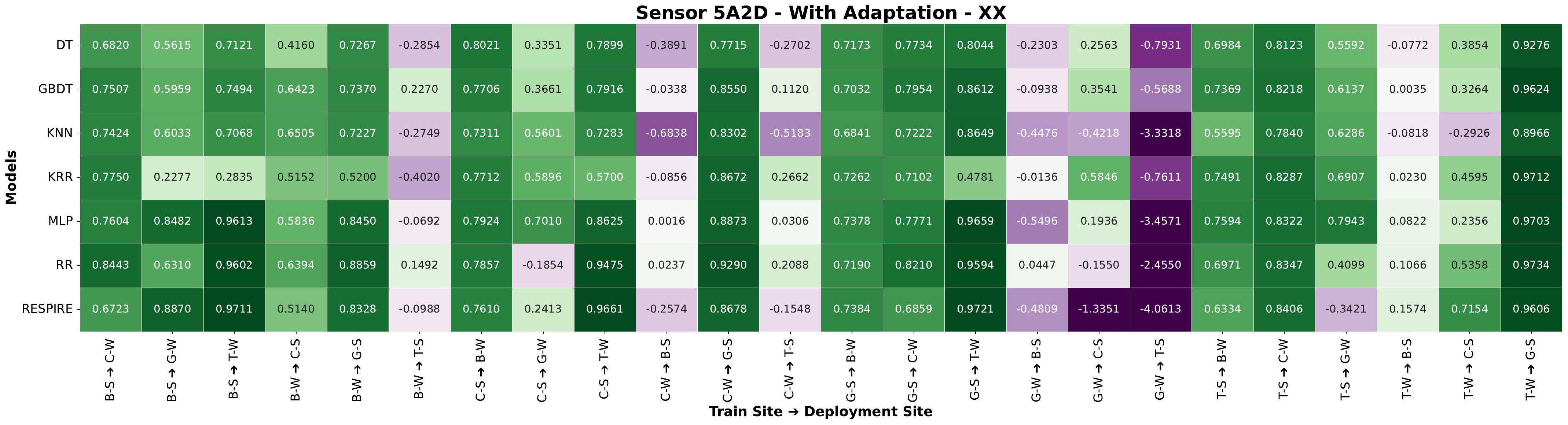}
        \end{minipage}%
    }
    \begin{minipage}{\textwidth}
        \centering
        \usebox{\spanningblock}%
        \raisebox{6.5em}{%
            \begin{minipage}[c][\ht\spanningblock][s]{0.06\textwidth}
                \centering
                \vfill
                \includegraphics[height=2.2\ht\spanningblock, width=\linewidth, keepaspectratio]{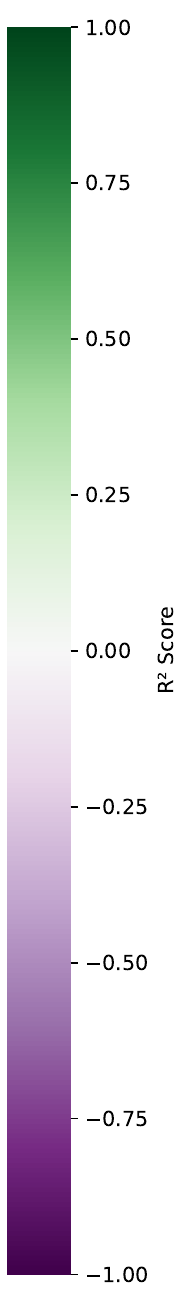}
                \vfill
            \end{minipage}%
        }
    \end{minipage}
    \caption{Heatmaps showing \metric performance of all baseline methods and \alg on training, non-transfer testing (SS) and transfer testing (SX, XS, XX). 1D adapters were offered to all methods. The same colorbar applies to all heatmaps with dark green corresponding to excellent performance \metric $\rightarrow 1$ and dark purple corresponding to poor performance \metric $\ll 0$.}
    \label{fig:main_label_part1}
\end{figure*}

\begin{figure*}[t]
    \centering
    \begin{subfigure}[b]{\textwidth}
        \centering
        \includegraphics[width=\textwidth]{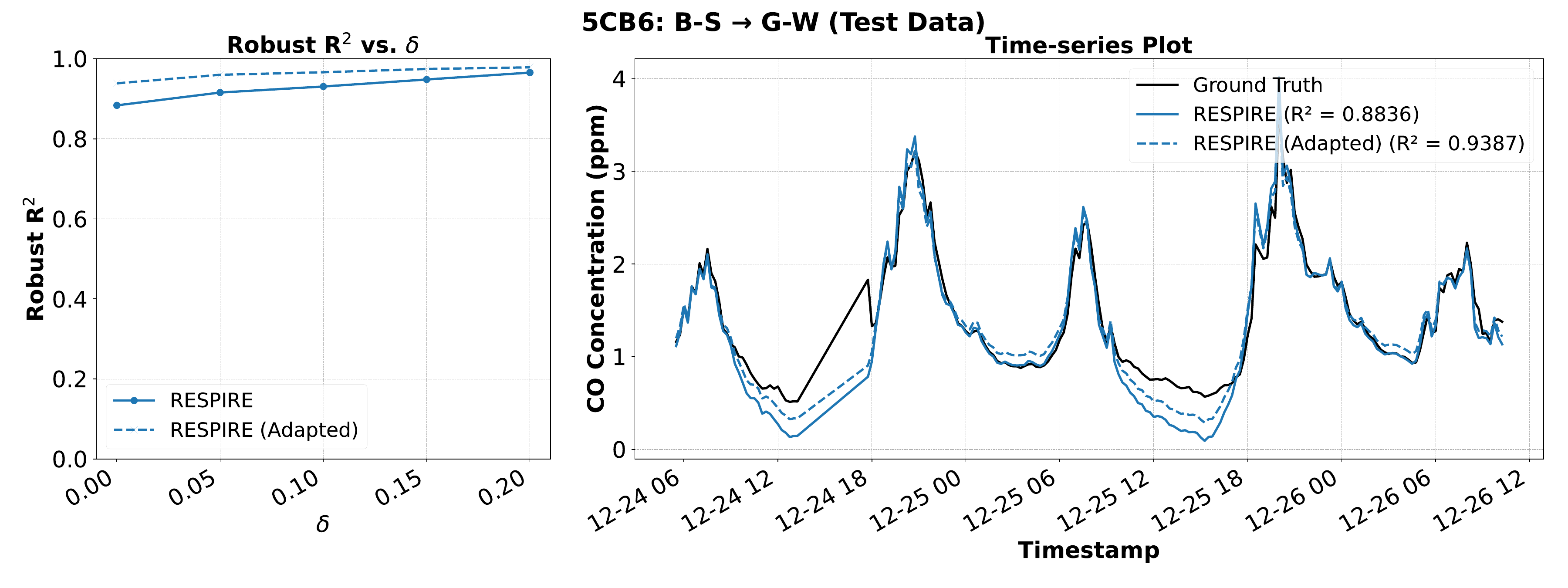}
    \end{subfigure}
    \hfill
    \begin{subfigure}[b]{\textwidth}
        \centering
        \includegraphics[width=\textwidth]{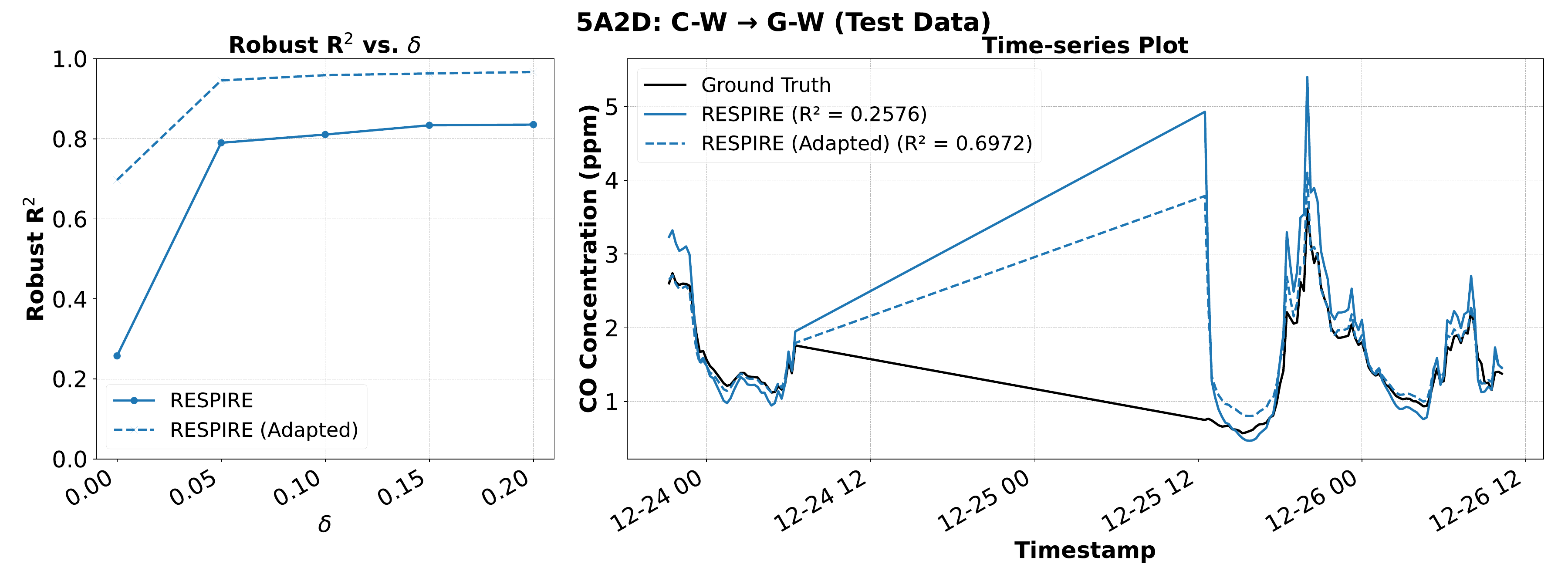}
    \end{subfigure}
    \caption{Two cases indicating the presence of outliers in data. Test \metric is often diminished due to a few points. The boxes on the left titled ``Robust \metric vs $\delta$" show how test \metric rapidly improves if just a $\delta$-fraction of the test data is excluded from evaluation. The boxes on the right show the corresponding time series of model predictions and ground truth. For the transfer C-W $\rightarrow$ G-W, a chunk of ground truth RGI data was missing and a very few bad predictions by the \alg model caused \metric to fall more than 50 percentage points from 0.8 to less than 0.25 (if no adapter is used) and almost 20 percentage points from 0.95 to 0.7 (if adapter is used).}
    \label{fig:row3}
\end{figure*}

\begin{figure*}[t]
    \centering

    \begin{minipage}[t]{\textwidth}
        \centering
        \begin{minipage}[b]{0.49\linewidth}
            \centering
            \includegraphics[width=\textwidth]{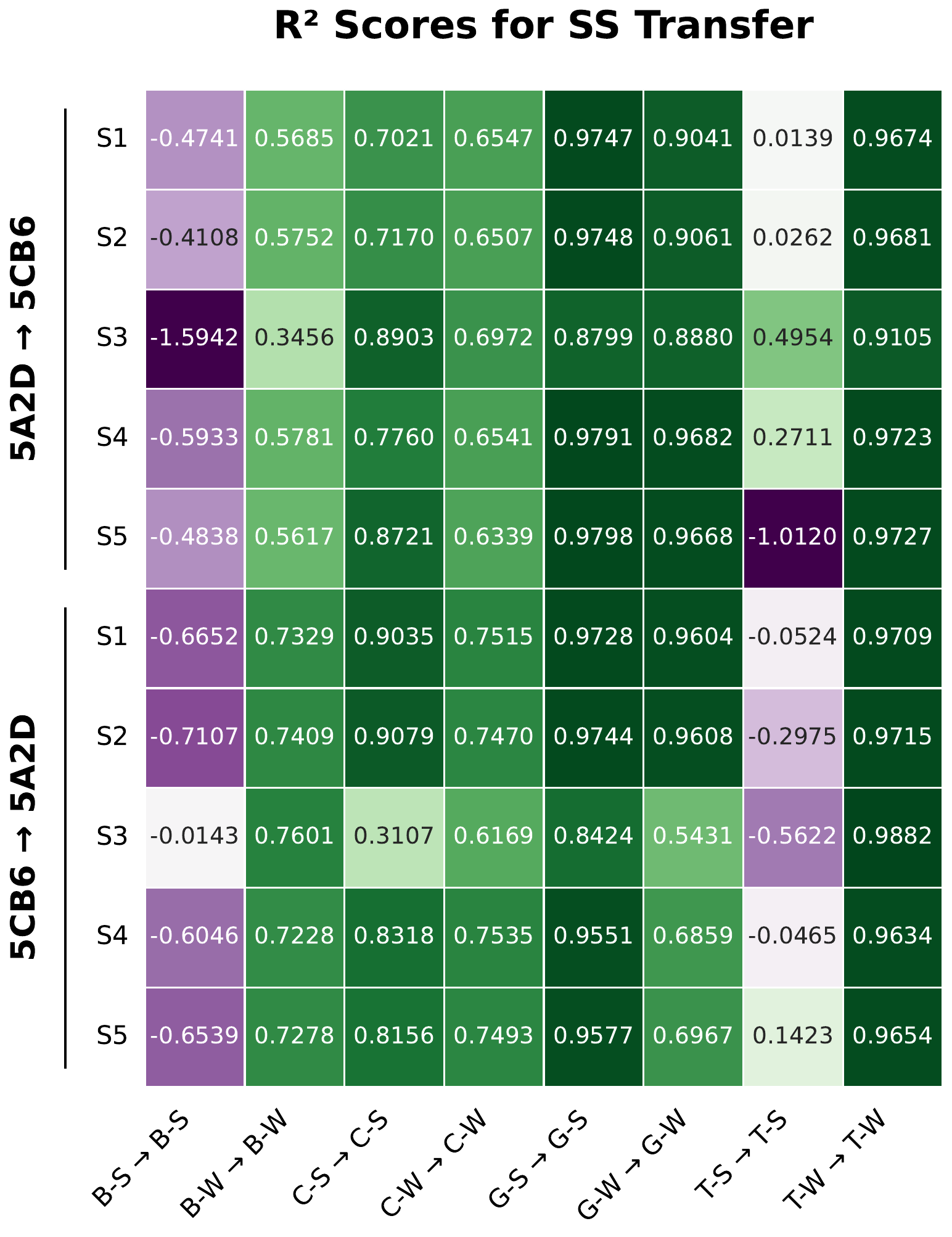}
        \end{minipage}\hfill
        \begin{minipage}[b]{0.49\linewidth}
            \centering
            \includegraphics[width=\textwidth]{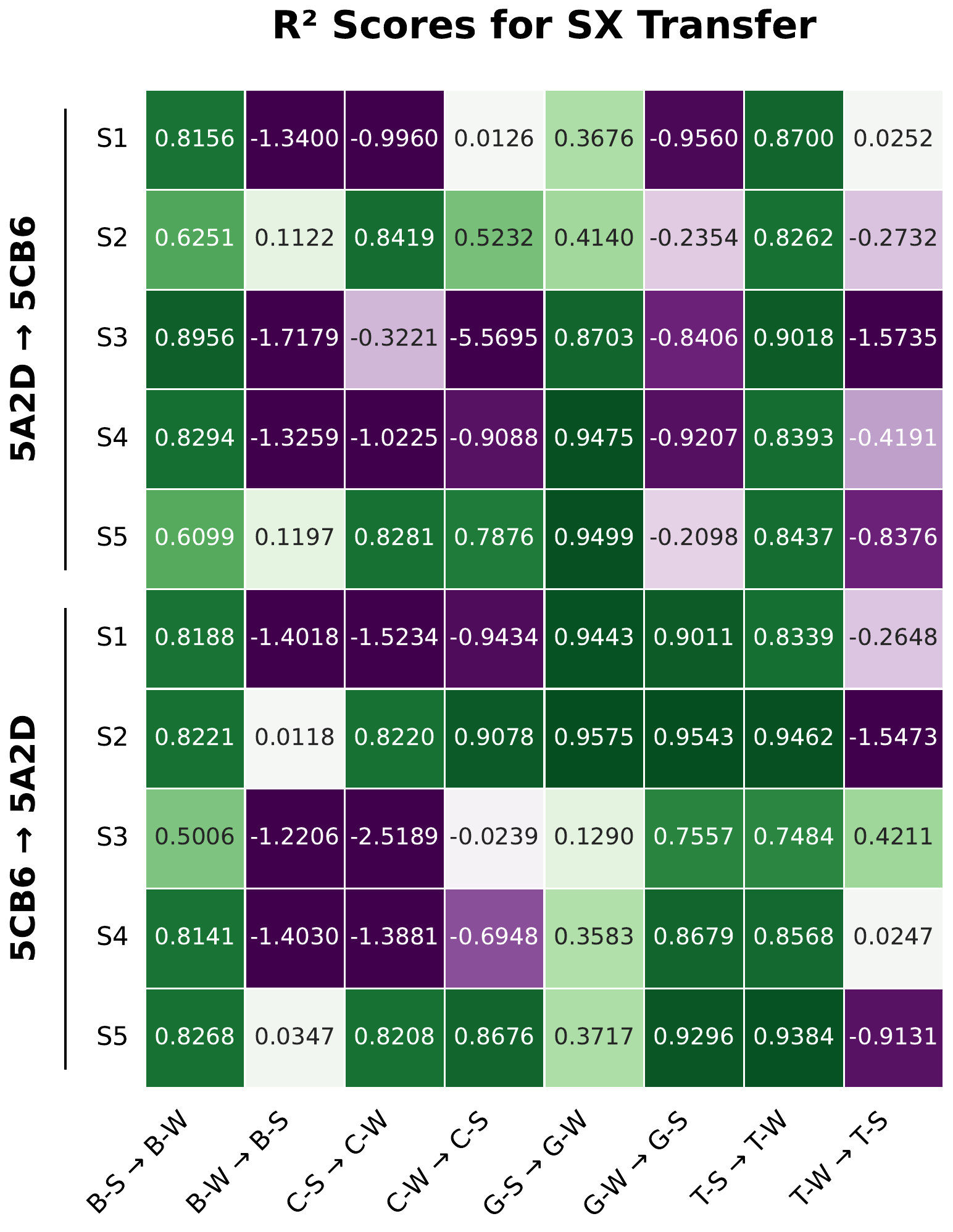}
        \end{minipage}
        
        \vspace{2mm}
        
        \captionof{figure}{Sensor-to-sensor transfer experiments for the pair of sensors 5A2D and 5CB6 (in both directions) for SS and SX transfer situations. As described in text, five scenarios are explored. S1 vs S3 results show that the \alg model can often be transferred without a transfer model being learnt. S3 vs S4 shows that learning a transfer model offers moderate benefits in some cases. S2 and S5 results correspond to S1 and S4 results, but with a 1D adapter learnt. The sensor-to-sensor adaptation exercise revealed interesting patterns about the sensors -- the median S5 \metric scores across all sensor-to-sensor transfers done from a sensor was positive for all the sensors except 5092 and D3EB. To investigate this further, Figure~\ref{fig:three_figures} shows KDE plots of the difference of S5 and S2 \metric scores after grouping the sensors into two groups: suspect (sensors 5092 and D3EB) and non-suspect (the rest of the sensors, except 1FEA since 1FEA had very few overlapping timestamps with other sensors).}
        \label{fig:s2s_transfer}
    \end{minipage}\hfill
    \raisebox{110pt}{%
    \begin{minipage}[t]{\textwidth}
        \centering
        \includegraphics[width=0.4\linewidth]{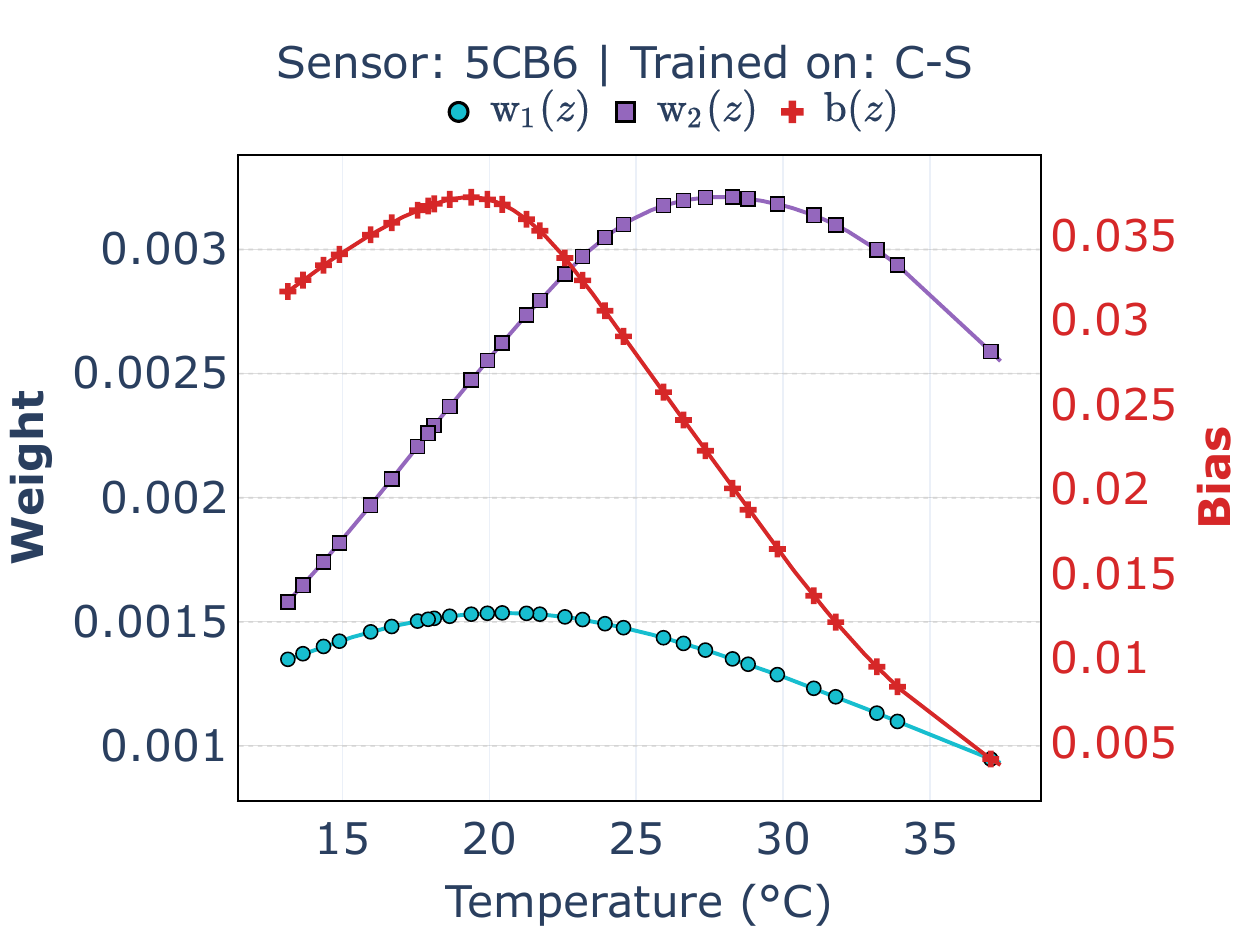}
        \includegraphics[width=0.4\linewidth]{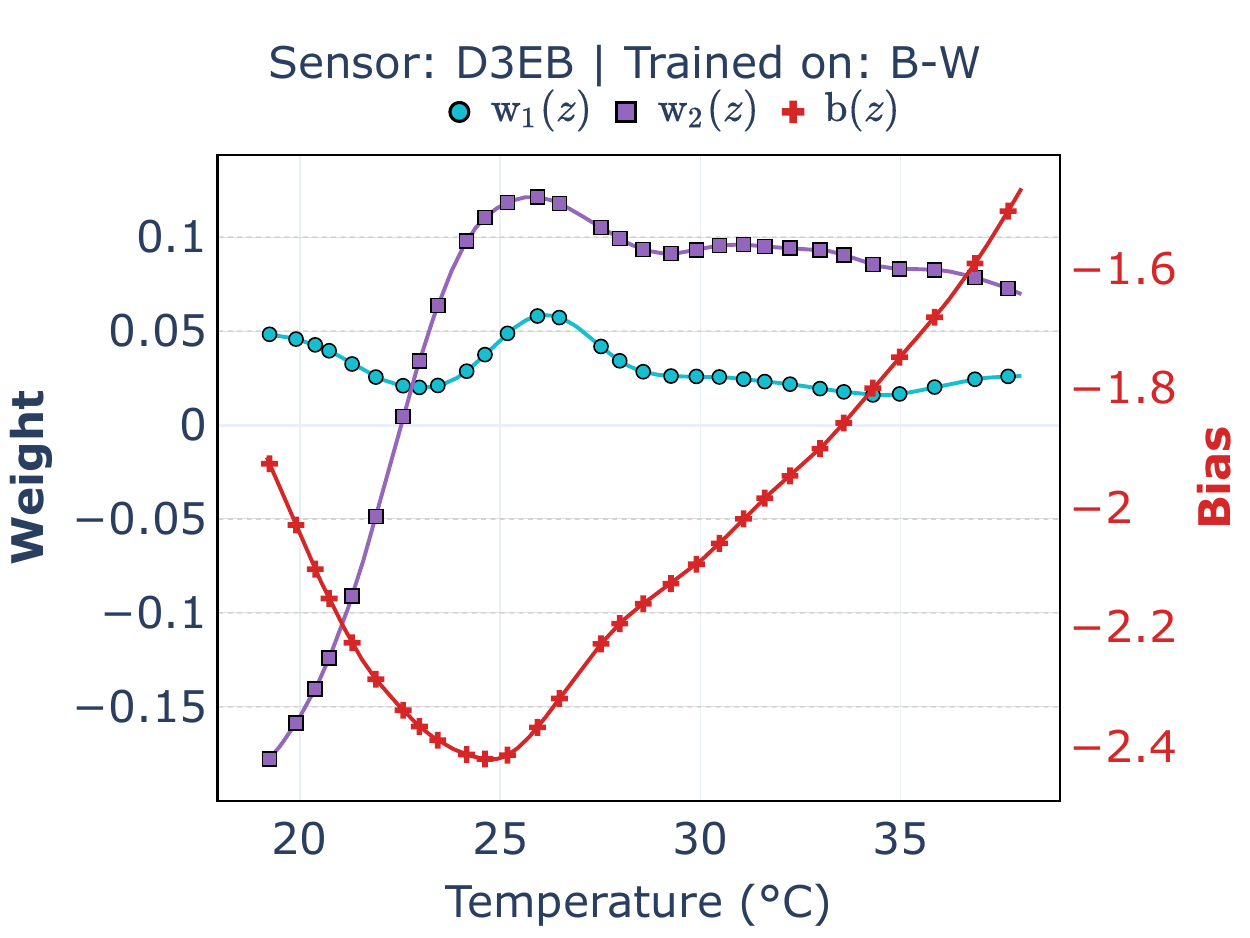}
        \captionof{figure}{\alg's semi-parametric model structure allows certain cases of overfitting to be detected. The figures above show the plot of weights $w_1, w_2$ and bias $b$ as a function of the auxiliary variable (ambient temperature). The model on the left hand side has smooth variations whereas the one on the right has larger and sharper variations. A paired two-sided t-test on the performance of these two models on various transfer scenarios confirmed the model on the left outperformed the model on the right with high confidence (p-value $\approx$ 0.001 in several cases).}
        \label{fig:row4}
    \end{minipage}%
    }

\end{figure*}

\begin{figure*}[t]
    \centering

    \begin{subfigure}{0.66\textwidth}
        \centering
        \includegraphics[width=\linewidth]{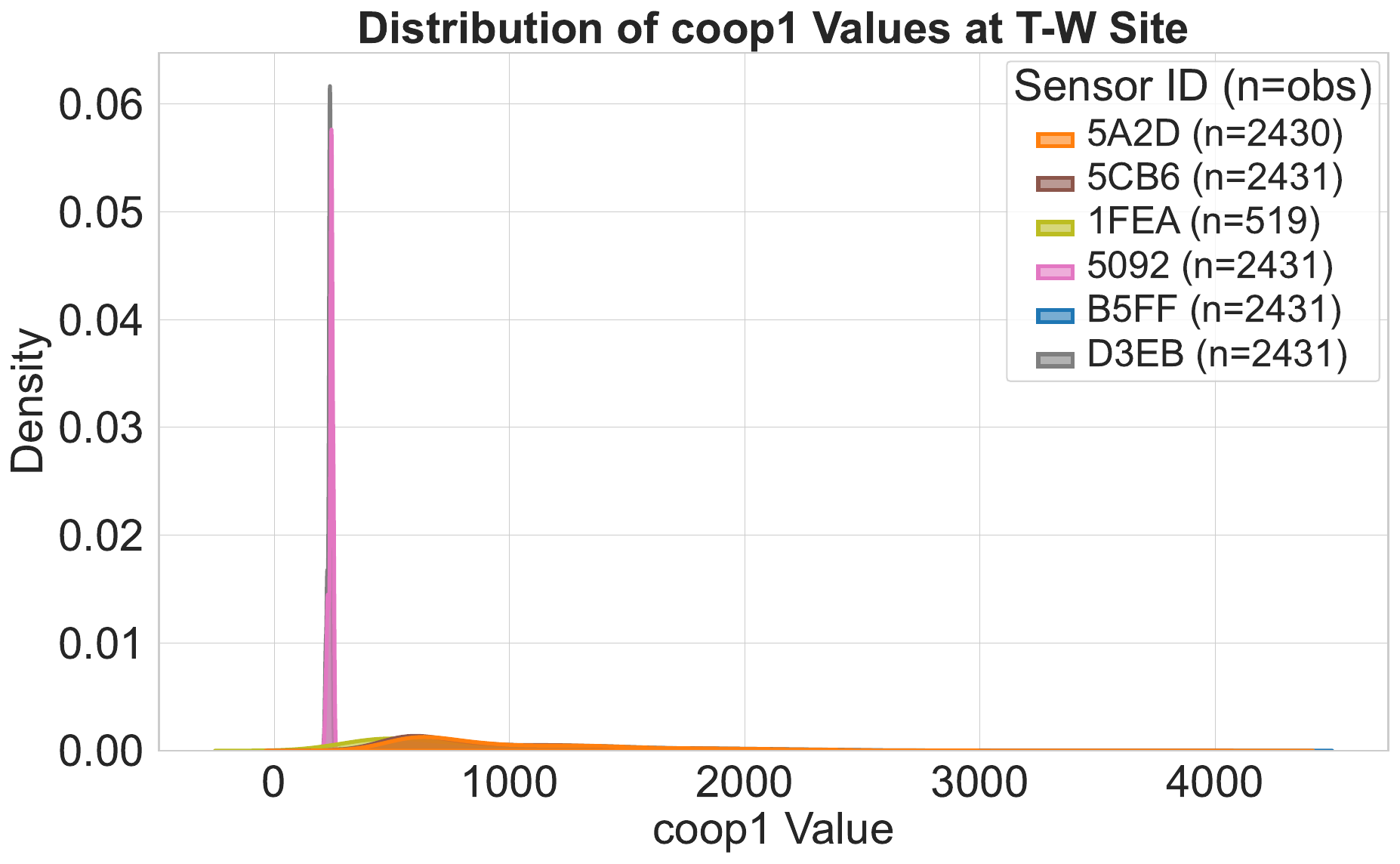}
    \end{subfigure}
    \hfill
    \begin{subfigure}{0.66\textwidth}
        \centering
        \includegraphics[width=\linewidth]{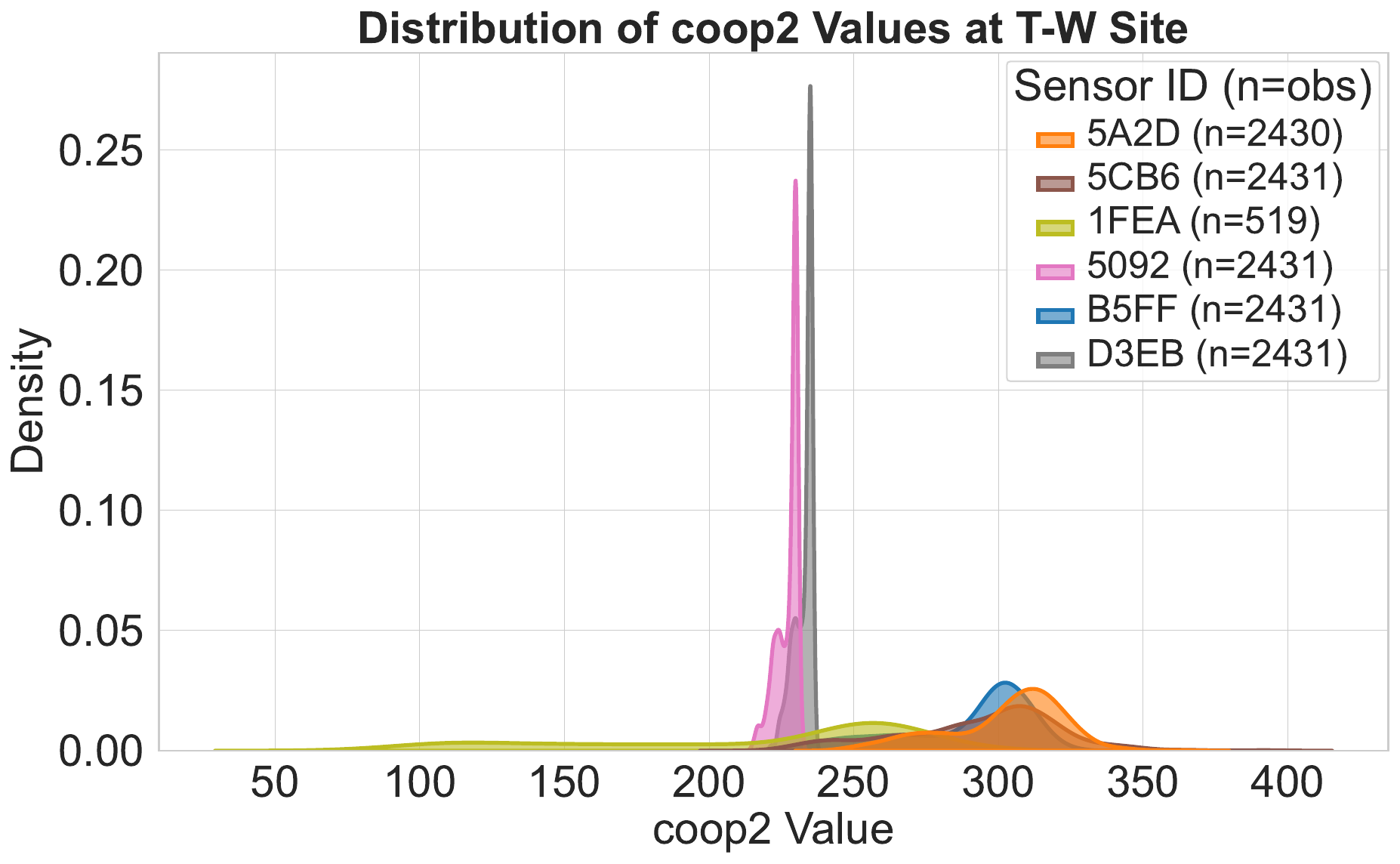}
    \end{subfigure}
    \hfill
    \begin{subfigure}{0.66\textwidth}
        \centering
        \includegraphics[width=\linewidth]{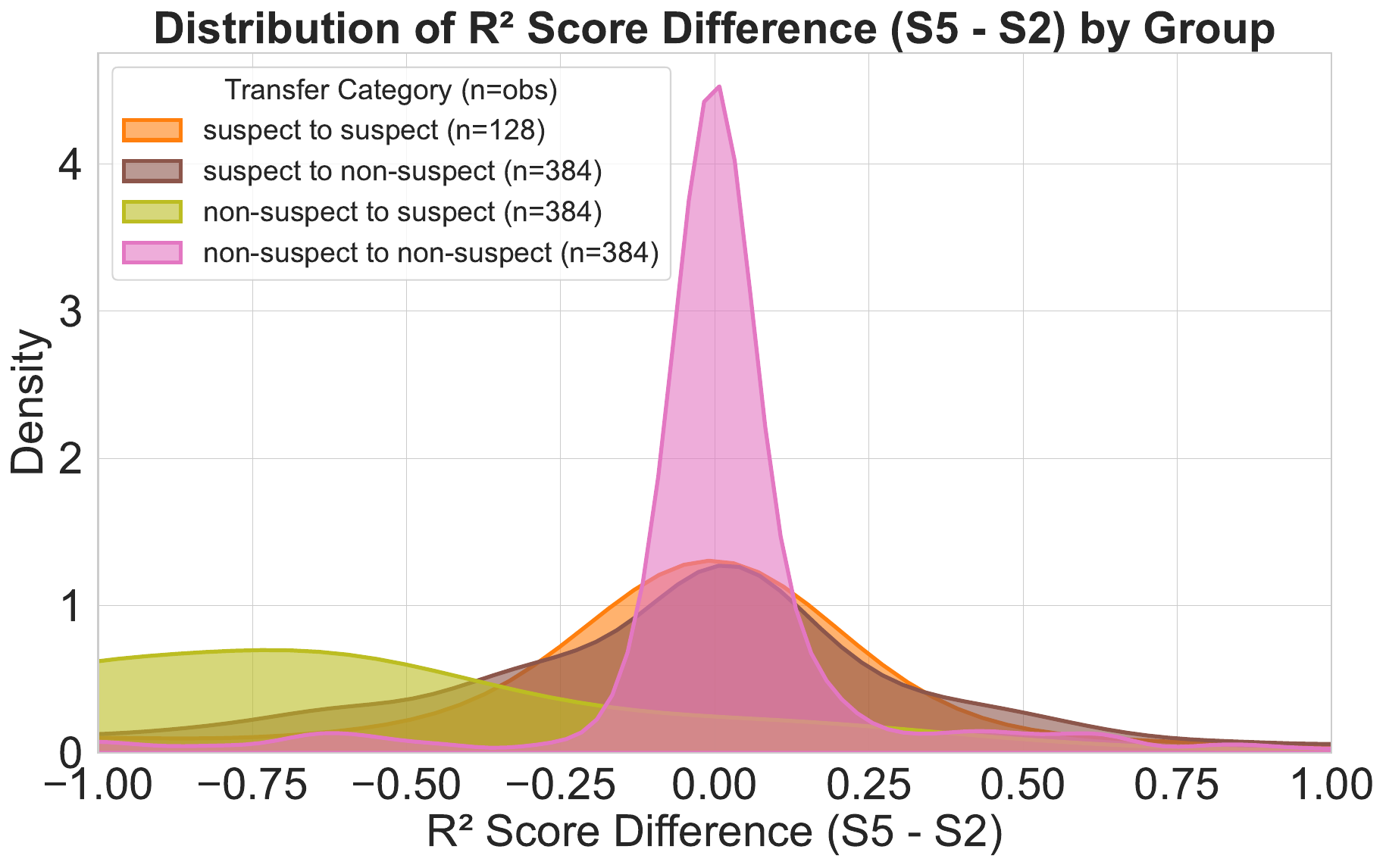}
    \end{subfigure}

    \caption{Results that indicate towards malfunction in the sensor packages 5092 and D3EB. The OP1 and OP2 values of these two sensors take significantly different values than the other sensors. KDE plots of S5-S2 \metric scores suggests that the models trained on the suspect sensors' data did not learn well in the first place and continue to perform poorly after sensor-to-sensor adaptation. The non-suspect sensors' models continue to perform well even after adaptation on non-suspect sensors. It is also notable that the \alg model overfit on data offered by sensor D3EB as reported in Figure~\ref{fig:row4}. This exercise was able to identify two suspect sensors for which it turned out (see Figures~\ref{fig:beforeafter} and \ref{fig:afterdist}), the readings for the \CO and \NOx sensors were swapped.}
    \label{fig:three_figures}
\end{figure*}

\begin{figure*}[t]
    \centering

    \begin{minipage}[t]{\textwidth}
        \centering
        \begin{minipage}[b]{0.49\linewidth}
            \centering
            \includegraphics[width=\textwidth]{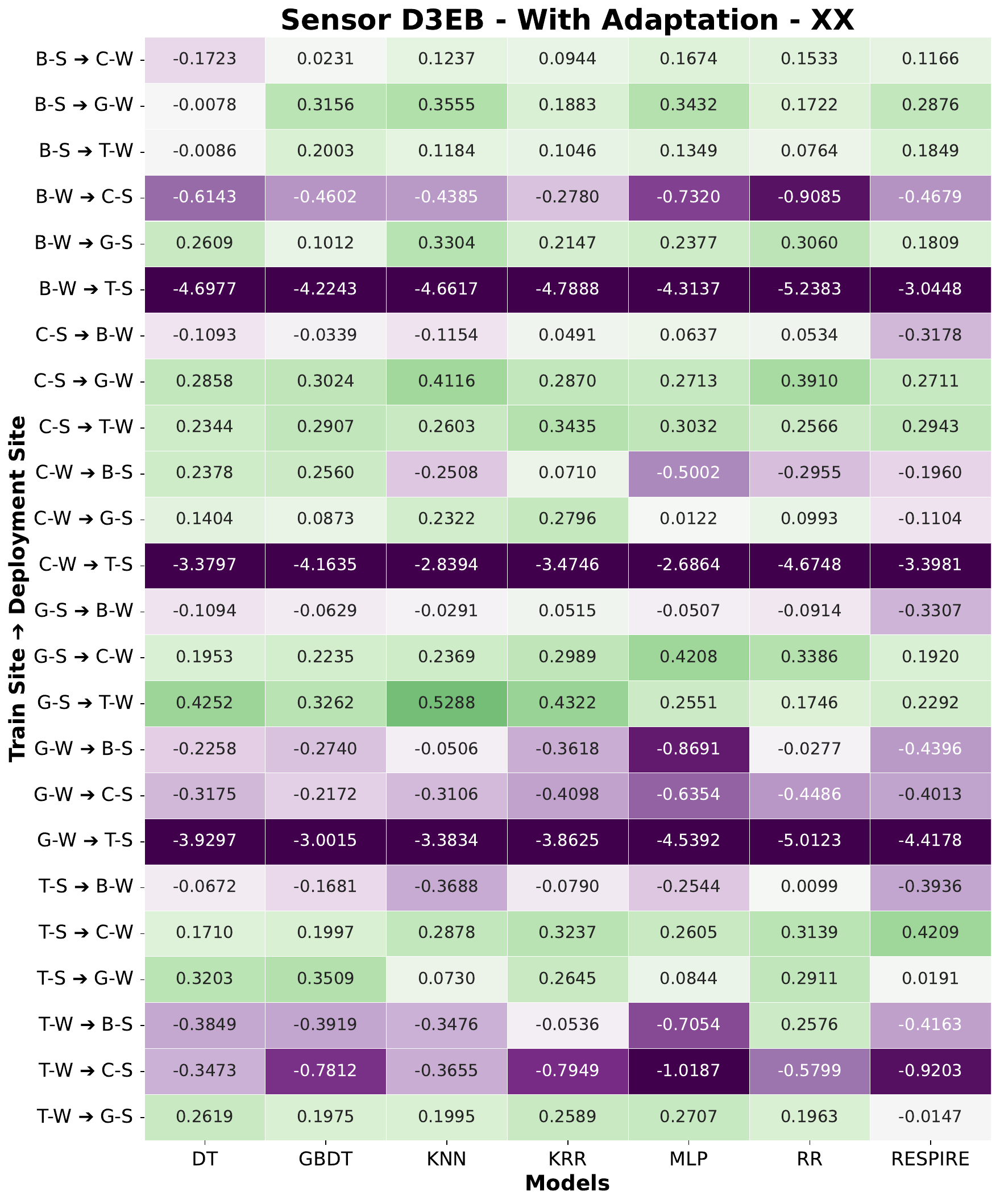}
        \end{minipage}\hfill
        \begin{minipage}[b]{0.49\linewidth}
            \centering
            \includegraphics[width=\textwidth]{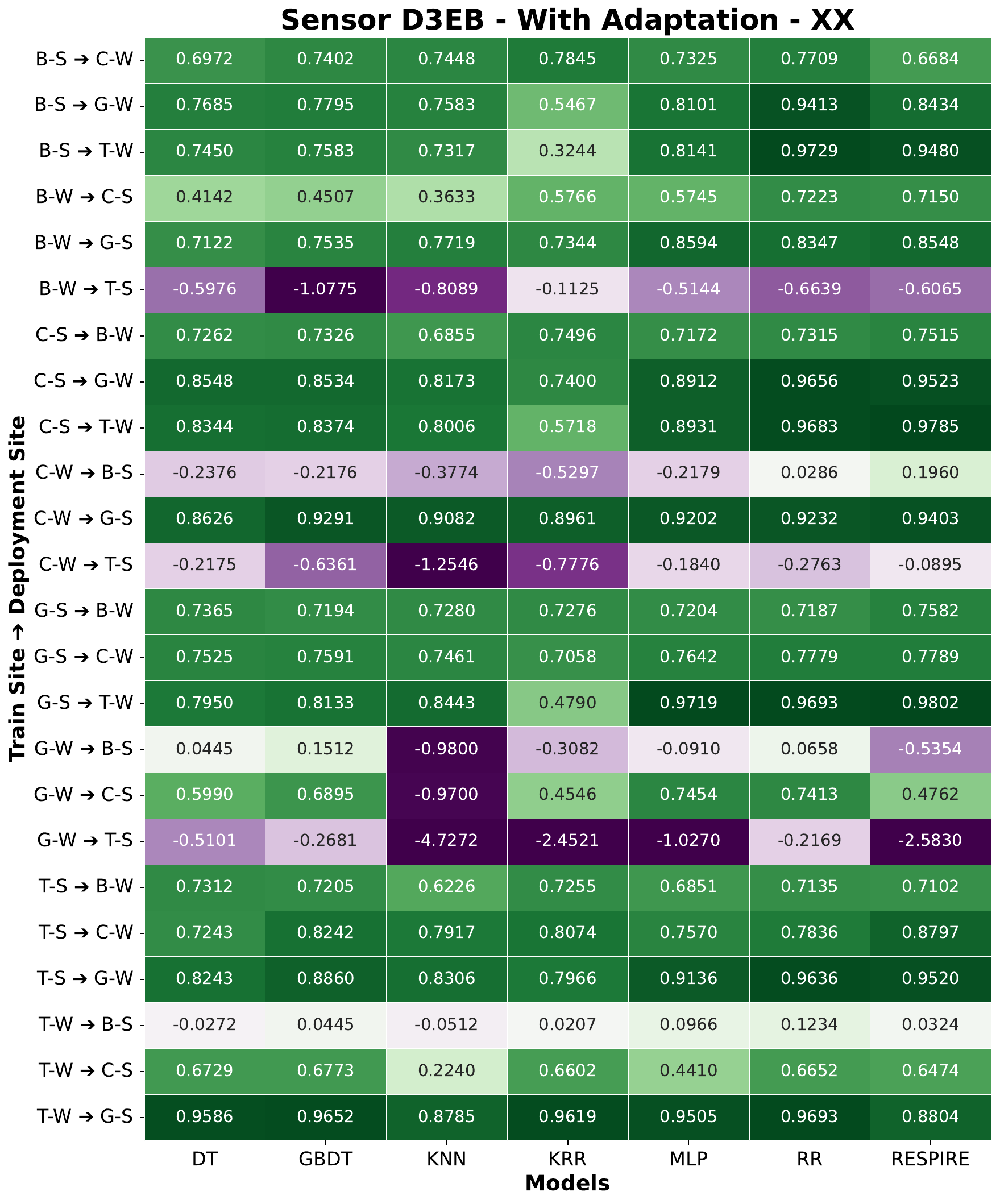}
        \end{minipage}
        
        \vspace{2mm}
        
        \captionof{figure}{On the left hand side are the \metric scores for the XX transfer experiment on data from the D3EB sensor before the sensor swap was corrected. On the right hand side are results after the swap was corrected. Correction of the \CO $\leftrightarrow$ \NOx swap led to significant improvement in performance across all methods and transfer experiments}
        \label{fig:beforeafter}
    \end{minipage}\hfill
    \raisebox{90pt}{%
    \begin{minipage}[t]{\textwidth}
        \centering
        \includegraphics[width=0.49\linewidth]{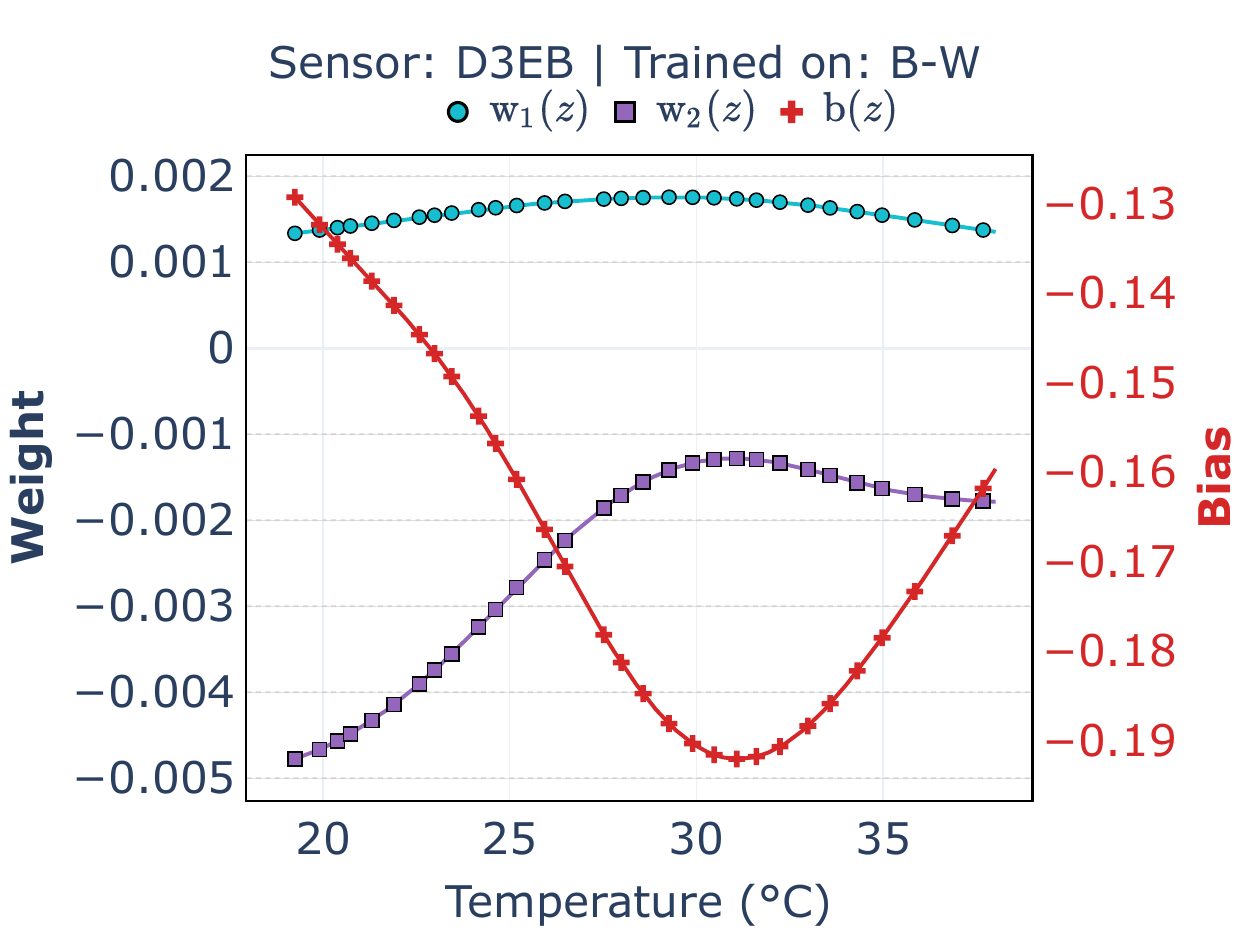}
        \includegraphics[width=0.49\linewidth]{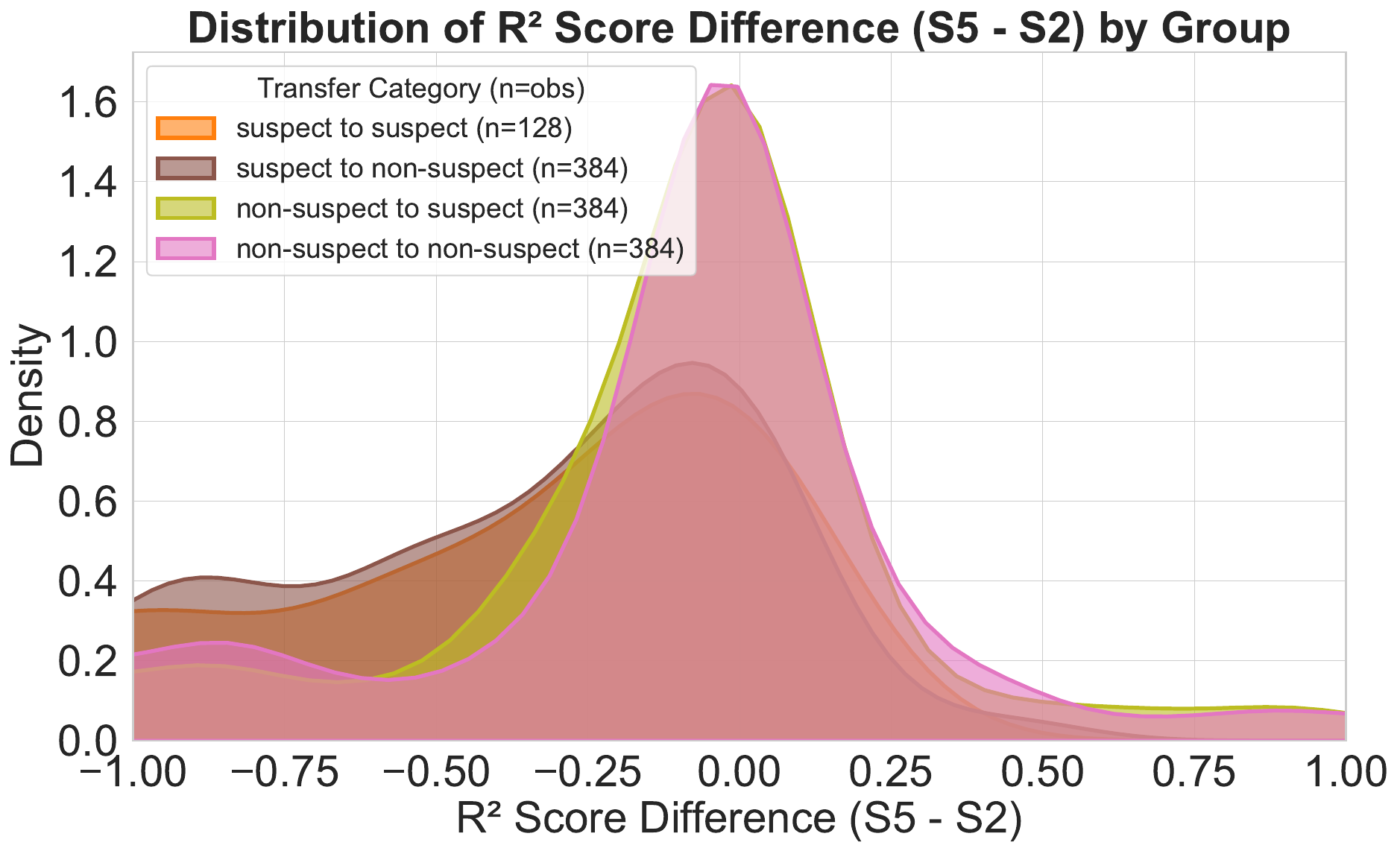}
        \captionof{figure}{Correcting the sensor swap led to significant improvement in the smoothness of the model parameters (as compared to Figure~\ref{fig:row4}) and the distributions of the S5-S2 scores (as compared to Figure~\ref{fig:three_figures}).}
        \label{fig:afterdist}
    \end{minipage}%
    }
\end{figure*}

\begin{figure*}[t]
    \centering
    \begin{subfigure}[b]{0.49\textwidth}
        \centering
        \includegraphics[width=\textwidth]{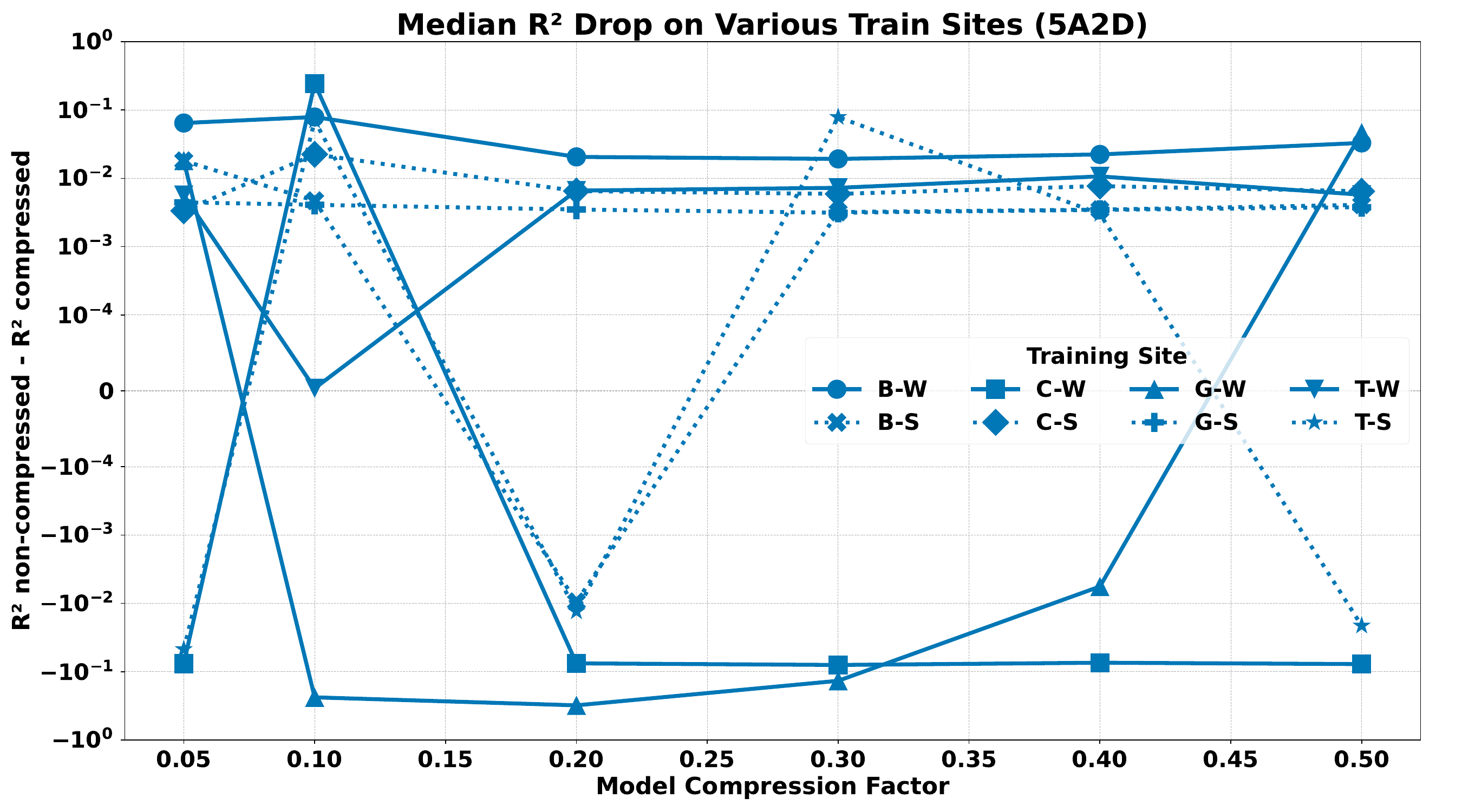}
    \end{subfigure}
    \hfill
    \begin{subfigure}[b]{0.49\textwidth}
        \centering
        \includegraphics[width=\textwidth]{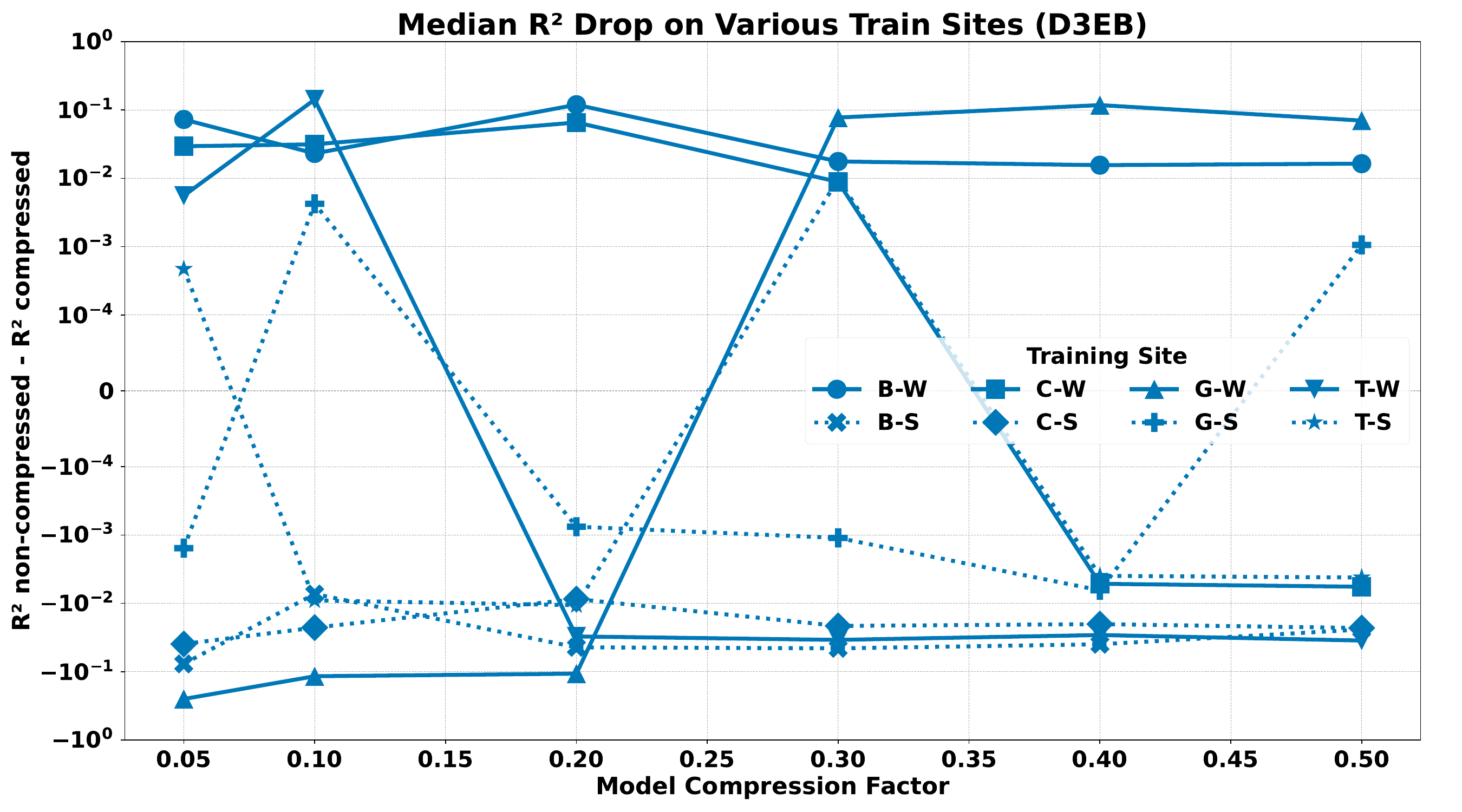}
    \end{subfigure}
    \caption{\alg models are compressible. The figures show negligible difference in prediction \metric scores even when the model is compressed more than 90\%.}
    \label{fig:compression_r2_drop}
\end{figure*}

\section{Empirical Results}
\label{sec:exps}

\mypar{Experimental Setup} For each sensor package and dataset, timestamp alignment was done. Only those timestamps were retained where the \CO sensor, temperature sensor and the \CO gas analyzer all had valid readings. For various datasets (see Table~\ref{tab:stats} for dataset details) approximately 500--2000 data points were available after carrying out this step. A temporal 80-20 split was created for each dataset by taking the first 80\% timestamps as the train portion and the last 20\% timestamps as the test portion. All experiments were carried out on a 64-bit machine with Intel® Core™ i7-6500U CPU @ 2.50GHz, 4 cores, 16 GB RAM and Ubuntu 24.04 OS. \alg code is available at \url{https://github.com/purushottamkar/respire}.\\

\mypar{Baseline Methods} Standard implementations from the popularly used \texttt{scikit-learn} library \cite{JMLR:v12:pedregosa11a} were used as baseline methods for empirical comparison. These included Regression Trees (\textbf{DT}), Gradient-boosted Decision Trees (\textbf{GBDT}), k-Nearest Neighbors (\textbf{KNN}), Kernel Ridge Regression (\textbf{KRR}), Multi-layered Perceptron (\textbf{MLP}), and Linear Ridge Regression (\textbf{RR}). The chosen baseline methods broadly cover the popular methods reported in prior literature for calibrating LCAQ \CO sensors \cite{levy,apostolopoulos,topalovic,zuidema,ariyaratne}.\\

\mypar{Hyperparameter Tuning} A range of hyperparameters was decided for each baseline method and \alg. The list of hyperparameter ranges offered to each method is described in Table~\ref{tab:hyperparameters}. For each experiment, a train dataset was chosen and for each and method participating in that experiment, hyperparameters were tuned using 3-fold cross-validation on the train portion of the dataset. The chosen hyperparameters were then used to train the final model on the entire train portion of that dataset. Bayesian optimization was used to accelerate hyperparameter search for \alg using a standard implementation from the \texttt{scikit-optimize} library \cite{tim_head_2020_3950294}. On the other hand, an exhaustive grid search was offered to baseline methods to tune their hyperparameters.\\

\mypar{Tuning of length Scale in the Matern Kernel} The selection of an appropriate \textbf{length scale} hyperparameter, $ls$, for the Matern kernel is crucial for achieving robust model generalization. In this work, we adopted a data-driven heuristic suggested by Garreau et. al. \cite{GarJitKan18} to define a relevant search space for $ls$. The procedure commences with the computation of the distribution of pairwise Euclidean distances, $d = \|x_i - x_j\|_2$, for all feature vectors $\{x_i, x_j\}$ in the training dataset. From this empirical distribution, a candidate set of length scales, $l_{cand}$, was derived by sampling at specific quantiles (e.g., 0.1, 0.25, 0.5, 0.75, 0.9), ensuring that hyperparameter search adapted to the scale of the data.\\

\mypar{Site-and-Season Transfer Experiments} To evaluate the ability of a model to perform on unseen sites and seasons, a variety of site-and-season transfer experiments were conducted. Recall from Section~\ref{sec:setting} that datasets are denoted by concatenating the site name and the season name, separated by a hyphen. Given this, four distinct transfers are studied:
\begin{enumerate}
    \item \textbf{SS} source and target datasets correspond to the same site and same season
    \item \textbf{SX} source and target datasets have the same site but different season
    \item \textbf{XS} source and target datasets have the same season but different sites
    \item \textbf{XX} source and target datasets have the different site and different season
\end{enumerate}

\mypar{Adapter Learning} In several cases, transfer experiments resulted in poor \metric scores simply due to predictions being off by a bias. This misrepresents the true performance of the model since the bias may have been introduced due to zero-shifts in the ground truth. To overcome this, a 1D adapter was trained on the model's prediction and the ground truth. In experiments where adaptation is done, an adapter is offered to all methods. It is notable that in all transfer experiments, training and hyperparameter tuning is done only on train dataset. If a 1D adapter is at all learnt, it is learnt only on the train part of the target dataset.\\

\mypar{Sensor-to-Sensor Transfer Experiments} In addition to transfers across sites and seasons, transfers across sensors were also considered. Given a pair of source and target sensors, say 5A2D $\rightarrow$ 5CB6 used in Figure~\ref{fig:s2s_transfer}, five distinct results are presented:
\begin{enumerate}
    \item (\textbf{S1}) the model's \metric on source sensor data without adaptation
    \item (\textbf{S2}) the model's \metric on source sensor data with adaptation
    \item (\textbf{S3}) the model's \metric on source target data without transfer model and without adaptation
    \item (\textbf{S4}) the model's \metric on source target data with transfer model but without adaptation
    \item (\textbf{S5}) the model's \metric on source target data with transfer model and with adaptation
\end{enumerate}
Transfer models, when learnt, were simply 2D-to-2D linear models trained to take operating potential (OP) values of the target sensor as input and predict the corresponding OP values for the source sensor as output. It is notable that this requires no supervision (no RGI \CO readings required).\\

\mypar{A Case of Swapped Sensors} \alg's semi-parametric model structure allows anomalous behavior, such as certain cases of overfitting, to be detected by inspecting the weights $w_1, w_2$ and bias $b$ predicted by the learnt model as a function of the auxiliary variable (ambient temperature). Figure~\ref{fig:row4} shows that these values demonstrate abnormal variations for the the sensor packages 5092 and D3EB. This indicated overfitting which was further confirmed by inspecting performance of the S5-S2 \metric scores as shown in Figure~\ref{fig:three_figures}. It turned out that in these two sensor packages, the readings for the \CO and \NOx sensors were swapped due to a firmware issue. Rectifying this issue immediately led to significant improvement in \metric scores for these two sensors (Figure~\ref{fig:beforeafter}) as well as the model weights $w_1, w_2$ and bias $b$ exhibited much less erratic behavior thereafter (Figure~\ref{fig:afterdist}). This chance exercise demonstrated the ability of semi-parametric models to offer not just explainable predictions, but also model weights with diagnostic value.

%% file: backmatter.tex
\section{Conclusion}
\label{sec:conc}
This paper presented results of a LCAQ sensor calibration exercise done on data from a large and diverse deployment of sensors and corresponding RGI setup in a mobile facility that toured four sites across several months. Some of the key takeaways of this study include the care needed to avoid overpowered, high-capacity models that may offer good same-site-same-season performance, but struggle if the site or season is switched.

The study presented favorable outcomes such as the success of sensor-to-sensor transfer of models that may allow large AQM networks to be established without each sensor requiring individual calibration. The challenges of cross-site and cross-season transfers were also discussed with zero-shifts that can cause models to give deceptively poor performance. Fortunately, in several cases, this can be mitigated by learning a simple 1D adapter. This paper also presented the \alg method for transferable calibration that closely aligns to manufacturer recommendations and offers the best performance in transfer experiments. The model also makes interpretable predictions with diagnostic value which, in two cases, allowed an inadvertent swap of sensor values to be detected and corrected.

Future work includes several directions such as experimenting with \alg for calibrating gas sensors of other types such as \Ox, \NOx. A more in-depth analysis of zero-shifts and drifts in both RGI and LCAQ data would be valuable to better understand the opportunities and limitations of current calibration approaches. Time series calibration techniques that accommodate meta data can also offer improved performance, such as time-of-day information for \Ox sensors.

\section*{Acknowledgment}
The work was supported by grant no 001296 from the Clean Air Fund. The authors are thankful to the members of the ATMAN collaboration for helpful discussions and to the members of the CSE lab staff team, especially Mr. Brajesh Kumar Mishra, Mr. Saurabh Malhotra, Mr. Saurabh Jaiswal and Mr. Karan Shah for timely and generous help with compute infrastructure and logistics. PK thanks Microsoft Research and Tower Research for research grants.

\bibliographystyle{unsrt}
\bibliography{refs}

\appendix

\input{appendix}

%% file: appendix.tex
\begin{table}[t]
    \centering
    \caption{Hyperparameter Search Space}
    \label{tab:hyperparameters}
    \resizebox{\textwidth}{!}{%
    \setlength{\tabcolsep}{4pt}%
    \begin{tabularx}{\textwidth}{
        >{\raggedright\arraybackslash}X 
        >{\raggedright\arraybackslash}X 
        >{\raggedright\arraybackslash}X 
        >{\raggedright\arraybackslash}X 
        >{\raggedright\arraybackslash}X 
        >{\raggedright\arraybackslash}X }
        \toprule
        \textbf{Hyper Parameter} & \textbf{Values / Range} & \textbf{Description} & \textbf{Hyper Parameter} & \textbf{Values / Range} & \textbf{Description} \\
        \midrule
        \multicolumn{3}{c}{\textbf{RESPIRE}} & \multicolumn{3}{c}{\textbf{Kernel Ridge (KRR)}} \\
        \midrule
        $\alpha$ & \{0, 0.05, 0.1, 0.15, 0.2\} & Proportion of outliers & $\lambda$ & \{0.1, 1, 10\} & L2 Penalty \\
        $q_{\text{ls}}$ & $[0.1,0.9]$ (continuous) & Quantile for kernel length- scale \cite{GarJitKan18} & $q_{\text{ls}}$ & \{0.1, 0.25, 0.5, 0.75, 0.9\} & Quantile for kernel length- scale \cite{GarJitKan18} \\
        $\eta$ & $[0.1,1.0]$ (continuous) & Outlier correction rate &  \\
        $\lambda$ & \{0.1, 0.5, 1, 5, 10\} & L2 Penalty & & & \\
        \midrule
        \multicolumn{3}{c}{\textbf{Decision Tree (DT)}} & \multicolumn{3}{c}{\textbf{MLP}} \\
        \midrule
        max\_depth & \{5, 10, 20, 40\} & Max tree depth & hidden\_layer
        \_sizes & \{(50,), (100,), (50,25)\} & Hidden-layer architecture \\
        min\_samples
        \_split & \{2, 5, 10\} & Samples to split & activation & \{relu, tanh\} & Activation function \\
        min\_samples
        \_leaf & \{1, 2, 4\} & Samples per leaf & $\lambda$ & \{1e-4, 1e-3, 1e-2\} & L2 Penalty \\
        \midrule
        \multicolumn{3}{c}{\textbf{Gradient Boosting (GBDT)}} & \multicolumn{3}{c}{\textbf{Ridge (RR) \& K-Nearest Neighbors (KNN)}} \\
        \midrule
        n\_estimators & \{50, 100, 200\} & Number of trees & {RR:} $\lambda$ & \{0.1, 1, 10, 50, 100\} & L2 Penalty \\
        \cmidrule(lr){4-6}
        learning\_rate & \{0.01, 0.1, 0.2\} & Shrinkage step & {KNN:} n\_neighbors & \{3, 5, 7, 10, 15\} & Size of neighborhood \\
        max\_depth & \{3, 5, 7\} & Tree depth & & & \\
        \bottomrule
    \end{tabularx}%
    }
\end{table}